\documentclass{article}
\usepackage{arxiv}
\usepackage{amsfonts}
\usepackage{amsmath}
\usepackage{amssymb}
\usepackage{amsthm}
\usepackage{bbm}
\usepackage{bm}
\usepackage{color}

\newcommand{\vect}[1]{\boldsymbol{#1}}
\usepackage[version=4]{mhchem}
\usepackage{subcaption}

\usepackage{graphicx}
\usepackage{physics}
\usepackage{systeme}
\usepackage{svg}
\usepackage{textgreek}
\usepackage{tikz}
\usetikzlibrary{positioning, fit}
\usepackage{amsfonts} 
\usepackage[normalem]{ulem}
\usepackage[utf8]{inputenc}
\usepackage{booktabs} 
\usepackage[vlined,ruled]{algorithm2e}

\SetKwInput{KwInput}{Input}                
\SetKwInput{KwOutput}{Output}              

\title{Double Diffusion Maps and their Latent Harmonics \\ for Scientific Computations in Latent Space }

\author{
  Nikolaos Evangelou\\
  Johns Hopkins University\\
  Baltimore, MD 21218, USA\\
  \And
  Felix Dietrich\\
  Technical University of Munich\\
  Munich, 80333, Germany\\
  \And
  Eliodoro Chiavazzo\\
  Polytechnic University of Turin\\
  Turin, 10129, Italy\\
  \And
  Daniel Lehmberg\\
  Technical University of Munich\\
  Munich, 80335, Germany\\
  \And
  Marina Meila\\
  University of Washington\\
  Seattle, WA 98195, USA\\
  \And
  Ioannis G. Kevrekidis*\\
 Johns Hopkins University\\
  Baltimore, MD 21218, USA\\
  yannisk@jhu.edu
}

\begin{document}
\maketitle
\begin{abstract}
We introduce a data-driven approach to building reduced dynamical models through manifold learning; the reduced latent space is discovered using Diffusion Maps (a manifold learning technique) on time series data. 
{\em A second round of Diffusion Maps on those latent coordinates} allows the approximation of the reduced dynamical models. 
This second round enables mapping the latent space coordinates back to the full ambient space (what is called \textit{lifting}); it also enables the approximation of full state functions of interest in terms of the reduced coordinates.
In our work, we develop and test three different reduced numerical simulation methodologies, either through pre-tabulation in the latent space and integration on the fly or by going \textit{back and forth} between the ambient space and the latent space. 
The data-driven latent space simulation results, based on the three different approaches, are validated through (a) the latent space observation of the full simulation through the Nystr\"om Extension formula, or through (b) lifting the reduced trajectory back to the full ambient space, via {\em Latent Harmonics}.
Latent space modeling often involves additional regularization to favor certain properties of the space over others, and  the mapping back to the ambient space is then constructed mostly independently from these properties; here, we use \textit{the same} data-driven approach to construct the latent space and then map back to the ambient space. 
\end{abstract}

\section{Introduction}
\label{sec:Introduction}

Differential equations (either ordinary or partial) are ubiquitous in science and engineering. They are often used to model spatiotemporal variations of quantities such as temperature, mass, energy, momentum, electric charge,  stresses and strains. Being able to efficiently (accurately and parsimoniously) solve differential equations underpins modeling in physics and engineering \cite{strogatz:2000}.
\begin{figure}[htb]
\begin{center}
\includegraphics[scale=0.36]{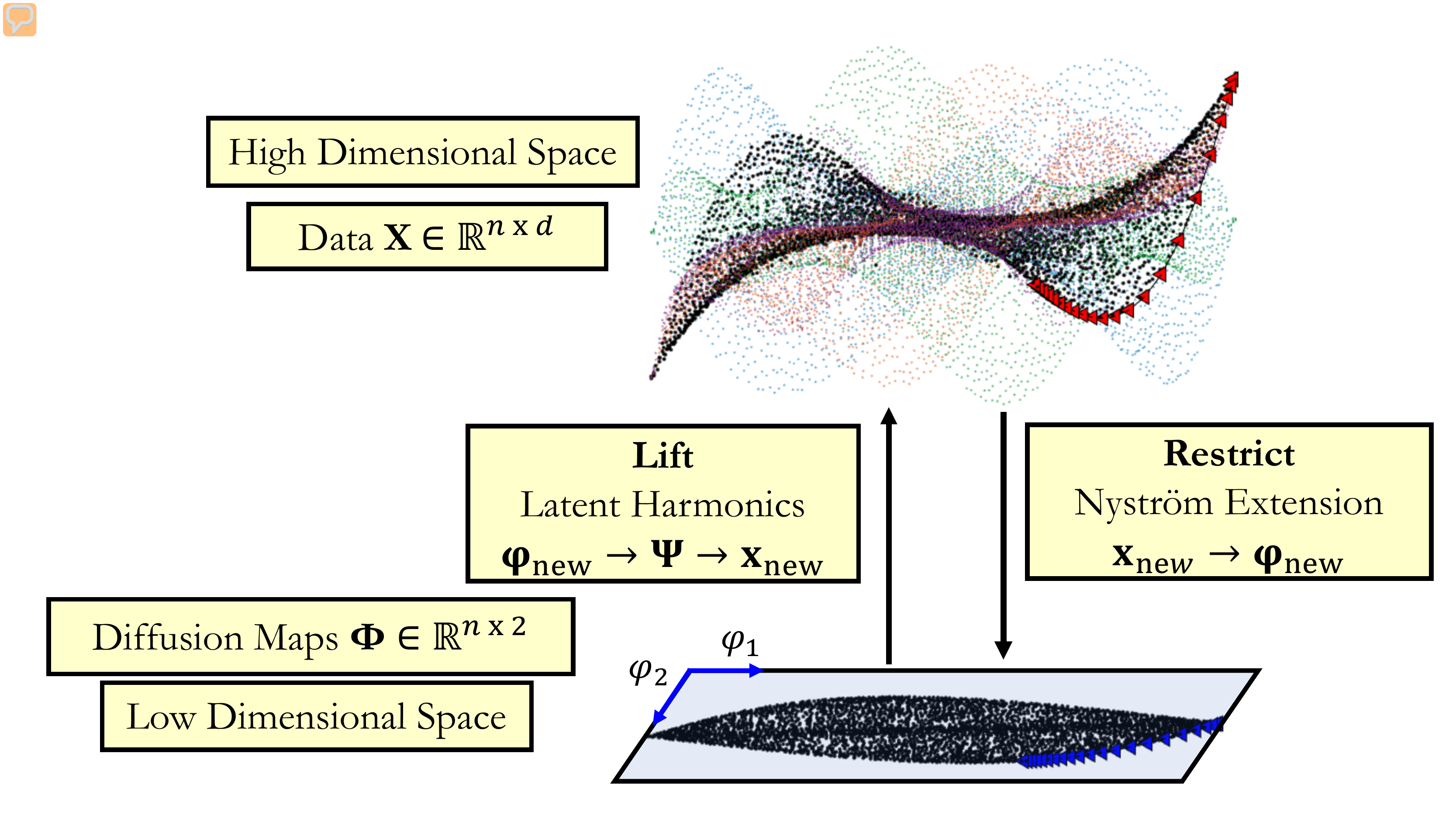}
\caption{An illustration of our computational tools (Diffusion Maps, Latent Harmonics and Nystr\"om Extension) for building reduced dynamical models.} 
\label{fig:Mapping_explanation}
\end{center}
\end{figure}

Numerically solving a time-dependent partial differential equation (PDE) relies on approximation through discretization techniques, such as finite differences, finite elements or spectral methods, which leads to finite dimensional dynamical systems \cite{Jolly_Pde_good_explanations}. This discretization ``transforms" the PDE to a system of Ordinary Differential Equations (ODEs). The more detailed tracking of the spatiotemporal evolution is required, the finer the spatial discretization needs to be and the larger the number of discretization ODEs that must be integrated \cite{Jolly_Pde_good_explanations}. 

Many PDEs arising in physicochemical modeling, however, exhibit long-time behavior that is relatively low-dimensional. 
From a mathematical point of view the long-term dynamics for those systems concentrate in an \textit{absorbing ball}, $\mathbb{B}_0$ and ultimately lie on a low-dimensional manifold \cite{Jolly_Pde_good_explanations,Inertial_Manifolds_Temam}.
For a class of such dissipative PDEs, including the Kuramoto-Sivashinsky equation (KSE) \cite{KSE_reaction_diff}, the Ginzburg Landau \cite{Ginzburg-Landau_LadayDoering_1988,Ginzburg-Landau_2}, Swift-Hohenberg \cite{Swift_Hohenberg} and several reaction-diffusion equations  \cite{KSE_reaction_diff,Reaction_Diffusion_Manif,sonday-2011,gear-2011,JollyInertialManifolds} an object called \textit{inertial manifold} exists \cite{Jolly_Pde_good_explanations}. 
Inertial manifolds are 
exponentially attracting, ``slow"  invariant manifolds that capture  the overall long-term behavior of the system \cite{sonday-2011}. Furthermore, restricting the dynamics to these slow inertial manifolds reduces dimensionality and mitigates stiffness of the initial, detailed discretized ODE system. This can significantly accelerate scientific computations.

Large stiff systems of ODEs arise also in other modeling contexts: control theory, where controlling forces may be turned in and off again suddenly, or circuit components having widely differing natural frequencies  \cite{Finite_Difference_Trefethen1996FiniteDA}. Another field in which stiff ODEs appear is chemical kinetics, in which the ordinary differential equations describe reactions of various chemical species to form other species. Here the stiffness is a consequence of the fact that different reactions take place on vastly different time scales (time scales ratios of $10^6$ or more are common). Due to the presence of fast and slow dynamics, the trajectories are initially attracted towards a slow manifold, and subsequently proceed to steady state, or to thermodynamic equilibrium, always remaining close to the manifold. The presence of such an attracting slow manifold  offers the possibility of  constructing a consistent reduced description of the process,
which accurately retains the slow dynamics along the manifold while neglecting the initial short transient approaching it in transverse directions \cite{Chiavazzo_2014,Finite_Difference_Trefethen1996FiniteDA}. 

Even when the existence of an \textit{inertial} manifold can be established, the theory usually does not provide of an explicit closed formula for it, and only approximations of the inertial manifold can be attempted \cite{Jolly_Pde_good_explanations}. 
Data Mining can be employed to discover such low dimensional inertial manifolds, given a collection of points sampled on it. Furthermore, machine learning algorithms can be employed to \textit{learn} the dynamics on the manifold without the need of closed form expressions.

This paper is organized  as follows: In Section \ref{sec:Double_DMaps} we describe the Double Diffusion Maps scheme. In Section \ref{sec:Algorithms} the proposed algorithms for the construction of reduced data-driven dynamical models are introduced. In Sections \ref{sec:Chafee_Infante} and \ref{sec:Combustion} we introduce our two examples: the Chafee-Infante PDE and a kinetic system comprised of stiff ODEs that models the combustion of hydrogen. In those Sections we discuss how sampling for these dynamical systems was performed and show the ability of Diffusion Maps to extract a useful set of latent coordinates, Algorithm \ref{alg:DMaps} and a more detail description in Section \ref{sec:Diffusion_Maps} of the Appendix. We further illustrate the ability of our data-driven scheme to learn the dynamics in the latent coordinates. We compare the data-driven trajectories with the solution computed by the equations in the ambient space. The comparison is made either by mapping the full trajectories in the reduced Diffusion Maps coordinates with Nystr\"om Extension (discussed in Section \ref{sec:Nystrom Extension} of the Appendix)  --what we call {\em Restriction}--  or by mapping the reduced data-driven trajectories back to the full space (\textit{lifting}) by using a regression scheme that operates in the latent coordinates that we call {\em Latent Harmonics}, described in Algorithm \ref{alg:DDMaps}. In Section \ref{sec:Conclusions}, we show that the latent coordinates obtained with the Diffusion Maps algorithm are one-to-one with the latent coordinates of an autoencoder. We also show the ability of two other regression schemes (Gaussian Processes and Neural Networks) to learn the dynamics in the Diffusion Maps coordinates. We conclude by summarizing the main advantages and disadvantages of our approach and discuss current research directions that we believe can improve the construction of accurate data-driven models.

We propose a data-driven framework for the construction of reduced data-driven surrogate models for dynamical systems whose long term dynamics are intrinsically lower dimensional. We combine the ability of the Diffusion Maps algorithm to (a) reduce the ambient dimension of a data set by discovering the intrinsic lower-dimensional geometry of the data manifold and (b) to extend functions defined on this manifold to perform scientific computations in latent space. We present a new use case of the Geometric Harmonics algorithm \cite{Geometric_harmonics_paper}, {\em Latent Harmonics},  that allow us to extend functions out of sample in the latent space.

We develop and test three multiscale numerical simulation methodologies by using tabulation in the low-dimensional space and integration on the fly or by mapping back and forth between the ambient space and the latent one, in the spirit of equation-free computing \cite{kevrekidis2003equation,kevrekidis2004equation,kevrekidis2009equation}
We also address the interpretability of our reduced models by mapping back to the ambient space (lifting) through the Latent Harmonics.

The construction of Reduced Order Models (ROMs) typically involves using a dimensionality reduction algorithm that discovers a set of latent coordinates, followed by learning  the dynamics in these coordinates. In the early 90s Proper Orthogonal Decomposition (POD) was used to find a set of reduced variables for chemical reacting systems or flows in complicated geometries \cite{deane1991low,krischer1993model,benner2015survey}. Learning the dynamics in terms of those reduced variables was done either with projection in the leading POD modes \cite{deane1991low} or by training a feed forward neural network \cite{krischer1993model}
for the evolution of the POD-Galerkin projection coefficients. Beyond these references,  POD has been proven useful in deriving reduced order models for a long list of applications in the literature \cite{ZHANG2003429,shvartsman2000order,shlizerman2012proper,FROUZAKIS200075}. 

However, since POD achieves dimensionality reduction by approximately passing hyperplanes through the data points, one might envision intrinsically low-dimensional {\em but curved} inertial manifolds that require much higher dimensional hyperplanes.  To circumvent this issue autoencoders have been used that project dynamical-systems onto non-linear manifolds \cite{lee2020model} and that simultaneously learn the governing equations and the reduced coordinates \cite{champion2019data}  by minimizing a combined loss functions.  
Our work deviates from Champion \textit{et al.} \cite{champion2019data} since no regularization is needed to discover the latent space coordinates and learn the dynamics. However, our method -even though based on harmonic analysis- retains a certain resemblence to autoencoders in that the same data-driven approach is used to both discover the latent space and then map back to the ambient space. In our case this is not an optimization problem (as it is for autoencoders), but rather a deterministic data-driven algorithm. 
Indeed, our group has used autoencoders to discover reduced latent spaces since the early 1990s \cite{rico1992discrete}; but at that time the nonlinear, noncnvex problem of training the autoencoder was much less straightforward than it is today, due to both software and hardware reasons. 

The work of Chiavazzo \textit{et al.} \cite{Chiavazzo_2014} and of Papaioannou\textit{et al.} \cite{papaioannou2021time} are perhaps the closest to ours. Chiavazzo \textit{et al.} use the example discussed also in Section \ref{sec:Combustion} from Chemical Kinetics for the construction of the reduced data-driven models. They mainly examine how different regression algorithms can be used to map from the Diffusion Maps coordinates back to the ambient space.
For the construction of the reduced data-driven models they use a similar approach to our Grid Tabulation (Section \ref{sec:Grid_Tabulation}) algorithm.
Papaioannou \textit{et al.} \cite{papaioannou2021time} used Diffusion Maps as one of the dimensionality reduction techniques to find a set of latent coordinates. They learn the dynamics in those coordinates with Multivariate Autoregressive (MVAR) and Gaussian Processes. Similar to our work they lift by using a version of Geometric Harmonics. The key difference between Chiavazzo \textit{et al.} \cite{Chiavazzo_2014} and Papaioannou \textit{et al.} \cite{papaioannou2021time}  and our current work is that, in order to lift with Geometric Harmonics, they use, at every new computational point, only the nearest few neighbors; we develop, implement and use a sensible global interpolation scheme here. 

\section{Double Diffusion Maps}
\label{sec:Double_DMaps}

The Diffusion Maps algorithm \cite{COIFMAN20065} can be used in two ways: (1) for dimensionality reduction of a finite data set, 
$ \mathbf{X} = \{ \vect{x}_i \}^{N}_{i=1} $ 
where $ \vect{x}_i \in \mathbb{R}^{m}$, sampled from a manifold,
$\mathcal{M}\subset \mathbb{R}^{m}$; or (2) to construct a function space basis called ``Geometric 
Harmonics'' that allows to extend functions defined on $\mathbf{X}, f(\mathbf{X}): \mathbf{X} \rightarrow \mathbb{R} $ \cite{Lafon-2004,COIFMAN20065,Geometric_harmonics_paper}. In this section, we describe the proposed Double Diffusion Maps scheme that employs two successive applications of Diffusion Maps.

\begin{figure}[h]
 \begin{center}
 \includegraphics[width=0.7\linewidth]{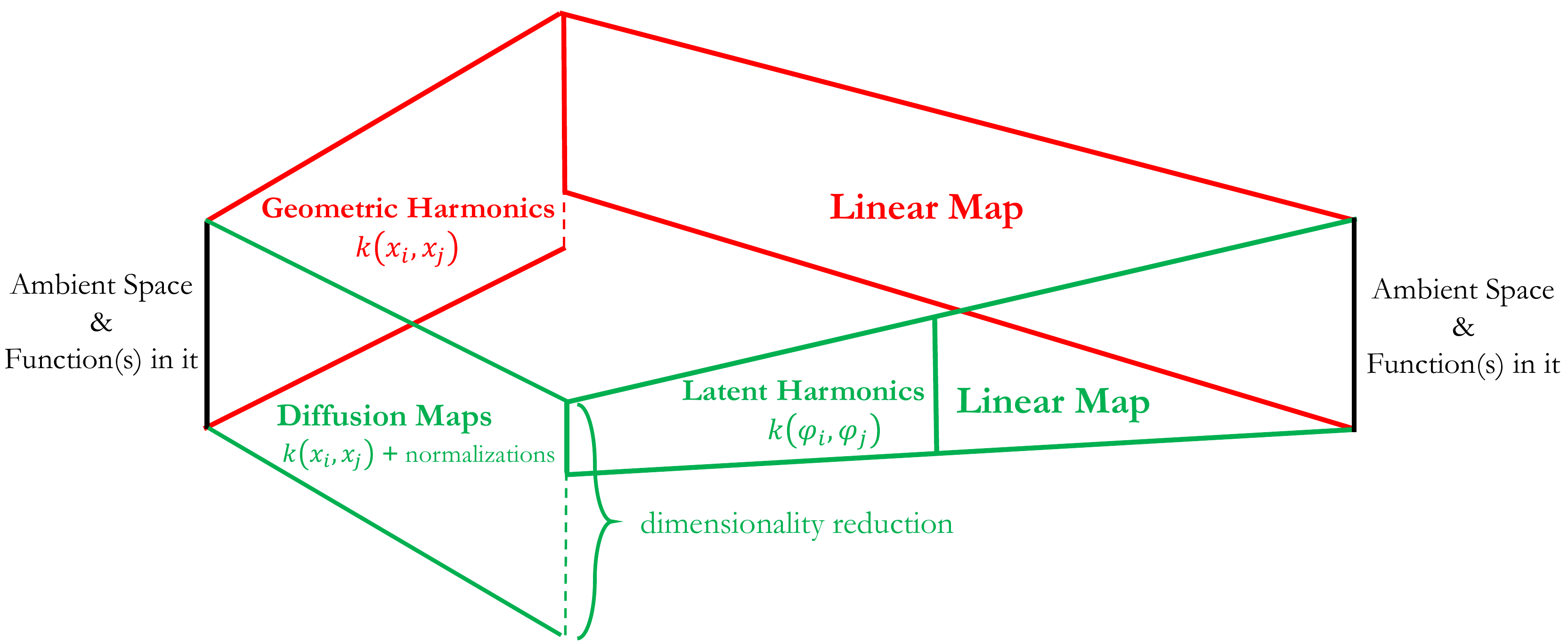}
\caption{The upper, red schematic shows the mapping constructed from the ambient space and functions in it, via Geometric Harmonics, back to the ambient space and functions in it.
The lower, green schematic illustrates our proposed approach that involves (a) first computing Diffusion Maps to discover the intrinsic geometry of a given data set (that might also lead to dimensionality reduction) and then (b) mapping from the {\em reduced} obtained latent space back to the ambient space and functions in it.}
 \label{fig:Old_New_Work}
 \end{center}
 \end{figure}

 Before we describe our Double Diffusion Maps Algorithm is worth mentioning briefly how the dimensionality reduction and the extension of a function can be achieved with Diffusion Maps.
The Diffusion Maps algorithm applied to a data set $\mathbf{X}$ provides a parametrization for it. This parametrization aims to reveal the intrinsic geometry of $\mathbf{X}$. If the number of the significant non-harmonic (see Section \ref{sec:Harmonics_Non_Harmonics} in the Appendix for a discussion of harmonic and non-harmonic eigenvectors) Diffusion Map coordinates is smaller than the dimension of the ambient space, the algorithm achieves dimensionality reduction.
Given a data set $\mathbf{X}$ sampled from a manifold $\mathcal{M}$, and a function $f$ defined on the sampled data set $\mathbf{X}$, Geometric Harmonics aims to extend $f$ for new points $\vect{x}_{new} \notin \mathbf{X}$. This map is constructed directly in the ambient space $\mathbb{R}^m$, Figure $\ref{fig:Old_New_Work}$.

In our suggested scheme we take advantage of the two use cases of Diffusion Maps (dimensionality reduction and extending a function), and show that it is possible to extend the function of interest $f$ by operating in only a few latent coordinates that parametrize the manifold, Figure \ref{fig:Old_New_Work}.
Our scheme proceeds as follows; first, we use Algorithm \ref{alg:DMaps} to obtain a few, reduced, non-harmonic coordinates $\mathbf{\Phi}$ (if $\mathbf{X}$ is lower dimensional); then we use Algorithm \ref{alg:DDMaps} to construct a function space basis that we call Latent Harmonics ($\mathbf{\Psi}$) by operating only on the leading non-harmonic eigenvectors, Figure \ref{fig:Old_New_Work}.

It is worth discussing why this second round of Diffusion Maps is indeed capable of extending functions defined on the original manifold. We remind the reader, that the eigenvectors $\mathbf{\Phi}$ were selected as the leading few non-harmonic eigenvectors that parametrize the (assumed low-dimensional) manifold; of course those were not the only computed eigenvectors resulting from the eigendecomposition in Algorithm $\ref{alg:DMaps}$. The discarded eigenvectors were functions of the ones selected, and therefore did not provide additional information in parametrizing the manifold. Yet, they are still important in extending a function using Geometric Harmonics - by discarding them we succeed in reducing dimensionality
but lose the ability of extending functions defined on the ambient data manifold.
In order to regain, our ability to accurately extend functions defined in the ambient data set
taking advantage of the intrinsic manifold low dimensionality, the reconstruction of the discarded harmonics becomes necessary. This reconstruction is achieved through Algorithm $\ref{alg:DDMaps}$ on $\mathbf{\Phi}$ through the computation of Latent Harmonics ($\mathbf{\Psi}$).

A particular example that we discuss in our numerical experiments is the extension of the state coordinates {\em in the high dimensional ambient space} for new points {\em in the latent space}, \hbox {$f$ : $\vect{\phi}_{new}$  $\rightarrow \vect{x}_{new}$}, what we refer to as \textit{Lifting}.

\begin{algorithm}[ht]
\caption{Dimensionality Reduction}
\label{alg:DMaps}
\SetKwFunction{FMain} {($\mathbf{X}$)}
  \DontPrintSemicolon
  \SetKwProg{Fn}{Diffusion Maps - Dimensionality Reduction}{:}{\KwRet}
  \Fn{\FMain}{
   1. For each $\{ \vect{x}_i\}_{i=1}^N  \in \mathbf{\Phi}$ where $\vect{x}_i \in \mathbb{R}^m$ compute $${K}(\vect{x}_i,\vect{x}_j) = \exp \left (-\frac{|| \vect{x}_i - \vect{x}_j||_2^2}{2\epsilon_1} \right ). $$ \\
   
  2. If $\mathbf{X}$ wasn't sampled uniformly: \\
  \hspace{2cm} (a) Compute $P_{ii} = \sum_{j=1}^N K_{ij}.$ \\
  \hspace{2cm} (b) Apply the normalization $\widetilde{\mathbf{K}} = \mathbf{P}^{-\alpha} \mathbf{K} \mathbf{P}^{-\alpha}$ where $\alpha=1$ to factor out the density. \\
  
  3. Normalize the transition matrix $\widetilde{\mathbf{K}}$, $$A(\vect{x}_i,\vect{x}_j)    = \frac{\widetilde{K}(\vect{x}_i,\vect{x}_j)}{\sum_{j=1}^N  \widetilde{K}(\vect{x}_i,\vect{x}_j) }.$$  \\
  3. Solve the eigendecomposition $\mathbf{A}\vect{\phi}_i = \lambda_i \vect{\phi}_i$ for each $i$ and then sort based on $\lambda_i$. \\
  
  4. Select the set of non-harmonic eigenvectors $\mathbf{\Phi} \in \mathbb{R}^{N \times k}$, where $k \leq m $ \cite{dsilva2015parsimonious}.

  \KwRet $\mathbf{\Phi}$;}
  \vspace{0.2cm}
  \end{algorithm}

\begin{algorithm}[ht]
\caption{Double Diffusion Maps  \& Latent Harmonics}
\label{alg:DDMaps}
\SetKwFunction{FMain} {($\vect{\Phi}$,$f$)}
  \DontPrintSemicolon
  \SetKwProg{Fn}{Double Diffusion Maps}{:}{\KwRet}
  \Fn{\FMain}{
   1. For each $\{ \vect{\phi}_i\}_{i=1}^N  \in \mathbf{\Phi}$ compute $${K^{\star}}(\vect{\phi}_i,\vect{\phi}_j) = \exp \left (-\frac{|| \vect{\phi}_i - \vect{\phi}_j||_2^2}{2\epsilon_2} \right ) $$
  
  2. Compute the $d$ first eigenvectors $\mathbf{\Psi} = \{\vect{\psi}_0, \dots,\vect{\psi}_{d-1}\}$ where $\vect{\psi}_i \in \mathbb{R}^{N}$ \\ and eigenvalues, $\vect{\sigma} = \{\sigma_0, \dots, \sigma_{d-1}\}$, of $\mathbf{K}^{\star}$  \\
  3. \textbf{Training}: Project $f$ on those eigenvectors
  $$ 
f \rightarrow P_{\delta}f = \sum_{j=1}^d \langle f,\vect{\psi}_{j}\rangle\vect{\psi}_{j,}
 $$
  \KwRet ;}
  \vspace{0.2cm}
  
  \SetKwFunction{FMain} {($\vect{\phi}_{new}$)}
  \DontPrintSemicolon
  \SetKwProg{Fn}{Latent Harmonics}{:}{\KwRet}
  \Fn{\FMain}{
   1. Compute the GH functions  for $\phi_{new}$ $$
    \Psi_{j}(\vect{\phi}_{new}) = \sigma^{-1}_{j}\sum_{i=1}^{m} {K^{\star}}(\vect{\phi}_{new},\vect{\phi}_i)\psi_{j}(\phi_i)$$
    where $\psi_j(\vect{\phi}_i)$ is the $i^{th}$ component of the $j^{th}$ eigenvector. \\
  2. \textbf{Extension}: Estimate the value of the function for $\phi_{new}$ $$
    (Ef)(\vect{\phi}_{new}) = \sum_{{j=1}}^d \langle f,\vect{\psi}_{j}\rangle\Psi_{j}(\vect{\phi}_{new})$$

  \KwRet $f(\vect{\phi}_{new})$;}
  
  \end{algorithm}

The computation of the latent coordinates discussed in Algorithm \ref{alg:DMaps} and the construction of the functional basis from Double Diffusion Maps Algorithm \ref{alg:DDMaps} need to be performed only once. 

Then the Latent Harmonics (Algorithm \ref{alg:DDMaps}) can be used to extend the values of any function originally defined in the ambient space for any $\vect{\phi}_{new} \notin \mathbf{\Phi}$ in the neighborhood of the training set.

In our work, we further exploit the \textit{Double Diffusion Maps-Latent Harmonics} scheme to perform scientific computations in the latent space. 

\section{Construction and Integration of Reduced Dynamical Models}
\label{sec:Algorithms}

In this section three different approaches are discussed for the construction and integration of reduced dynamical models. For the described algorithms below, it is assumed that the latent Diffusion Maps coordinates in which the models are constructed have already been obtained, Algorithm \ref{alg:DMaps}. 
To construct and validate those reduced dynamical models, global interpolation schemes are important that allows us to map new unobserved data from the ambient space coordinates to the reduced space original variables, restriction, but also from the reduced coordinates back to the original ones, lifting. In our work we \textit{restrict} by using the Nystr\"om Extension, described in Section \ref{sec:Nystrom Extension} of the Appendix. The inverse mapping of \textit{lifting} is achieved by performing Algorithm \ref{alg:DDMaps}. In general the Latent Harmonics allows to construct any smooth function of interest defined in the ambient or reduced space in terms of the few parsimonious reduced Diffusion Maps coordinates in which we aim to construct the data-driven surrogate model.

The aforementioned scientific tools are important to construct reduced models in terms of the leading non-harmonic Diffusion Maps coordinates. Of course building dynamical models in the Diffusion Maps coordinates means that we should be able to compute the evolution of $\vect{\phi}(t)$ with time. Expressing the evolution of $\vect{\phi}(t)$ through a simple numerical scheme such as Forward Euler we have:
\begin{equation}
    \label{Euler}
    \vect{\phi}_{t_{i+1}} = \vect{\phi}_{t_{i}} + h\left (\frac{d\vect{\phi}_{t_{i}}}{dt} \right )
\end{equation}
 where $\vect{\phi}_{t_{i}} =  \vect{\phi} (t = i)$ and $h$ is the step size. The computation of the dynamics in terms of $\vect{\phi}$ is non trivial, since we do not have access to closed form expressions of the right hand side of the evolution equations in the low dimensional space. It becomes possible to compute $\frac{d\vect{\phi}_{t_{i}}}{dt}$ since differentiating $\vect{\phi}(\vect{x}(t)) $    
 through the chain rule gives
\begin{equation}
    \label{eq:chain rule}
    \frac{d\vect{\phi}_{t_i}}{dt} = \frac{d\vect{\phi}_{t_i}}{d\vect{x}_{t_i}}\cdot \frac{d\vect{x}_{t_i}}{dt},
\end{equation}
where $\frac{d\vect{x}_{t_i}}{dt}$ are the known dynamics in the ambient space coordinates for which we have analytical expressions. The term $\frac{d\vect{\phi}_{t_i}}{d\vect{x}_{t_i}}$ is computed by symbolic differentiation of the Nystr\"om Extension Formula described in Section \ref{sec:Nystrom Extension} of the Appendix. These derivatives could be also computed by ``automatic differentiation'' that nowadays is available  for scientific computations even when the function is only available in the form of a black box.

 In our described algorithms below we show how the reduced dynamics are estimated by repeatedly going \textit{Back and Forth} (BF) between the ambient space coordinates and the Diffusion Maps coordinates, Section \ref{sec:Back-Forth}. Connections between this BF algorithm and the Equation-Free modeling approach of \cite{kevrekidis2003equation} are described in Section \ref{sec:Equation_Free} the Appendix. A second algorithm, described in Section \ref{sec:Grid_Tabulation} (GT) has many similarities with \cite{Chiavazzo_2014} in which the reduced dynamics are estimated and tabulated in the nodes of a Cartesian grid.  
 The difference between our approach and the one discussed in \cite{Chiavazzo_2014} has to do with the computation of the reduced vector field on the nodes of the grid; further details are given in Section \ref{sec:Grid_Tabulation}. 
 In our last algorithm \textit{Tabulation with Latent Harmonics Interpolation} (TaLHI), described in Section \ref{sec:DDMAps Tabulation} $\frac{d\phi}{dt}$ is computed for {\em all the collected data points} $\vect{x}_i \in \mathbf{X}$ and globally tabulated.

 \subsection{Back and Forth}
\label{sec:Back-Forth}

  The BF approach allow us to integrate in the reduced coordinates, $\vect{\Phi}$, by repeatedly going Back and Forth between the reduced space and the ambient space. Algorithm \ref{alg:Back_and_Forth} below assumes as input a point in the reduced space. If an initial condition in the ambient space is given,  the corresponding value in the Diffusion Maps coordinates can be computed through the Nystr\"om Extension formula, Section \ref{sec:Nystrom Extension}. \vspace{5mm}

 \begin{algorithm}[ht]
 \caption{Back and Forth}
 \label{alg:Back_and_Forth}
 \SetKwFunction{FMain} {($\vect{\phi}_{t_i}$)}

   \DontPrintSemicolon
   \SetKwProg{Fn}{Input}{:}{\KwRet}
   \Fn{\FMain}{
   
   1. Use the extension scheme from Algorithm \ref{alg:DDMaps} to \textit{lift} 
   $\vect{\phi}_{t_i}$. \\
   2. Compute $\frac{d\vect{x}_{t_i}}{dt} = \vect{f}(\vect{x}_{t_i})$ with the original model.\\ 
   3. Compute $\frac{d\vect{\phi}_{t_i}}{dt}$ from the chain rule in Equation \ref{eq:chain rule}. 
   
  \KwRet $\frac{d\vect{\phi}_{t_i}}{dt}$\;}
   \end{algorithm}

\begin{figure}[ht]
 \begin{center}
 \includegraphics[width=0.6\linewidth]{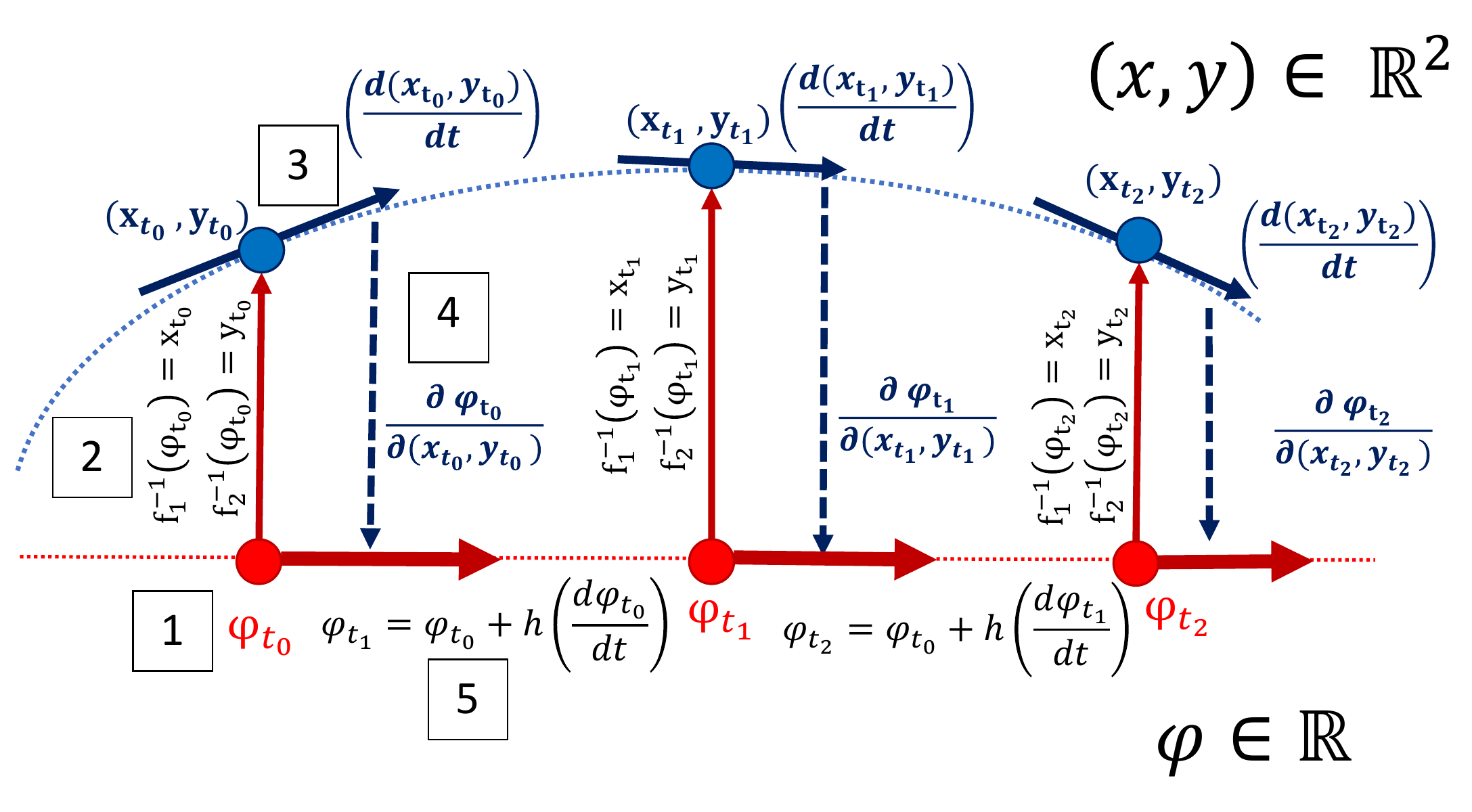}
\caption
{1. Start integration at $t = 0$ from an initial condition in the reduced space $\phi_{t_{0}}$. 2. Use Algorithm \ref{alg:DDMaps} to compute the corresponding values of the high dimensional space variables, $x_{t_{0}},y_{t_{0}}$. 3. From  $x_{t_{0}},y_{t_{0}}$ by using the expressions of the dynamics in the high dimensional space compute the derivatives $\frac{d(x_{t_{0}},y_{t_{0}})}{dt}$. 4. To obtain the reduced dynamics $\frac{d\phi_{t_{0}}}{dt}$ the calculation of the chain rule is needed $\frac{d\phi_{t_0}}{dt} =  \frac{\partial\phi_{t_{0}}}{\partial x_{t_{0}}} \cdot \frac{dx_{t_{0}}}{dt} + \frac{\partial\phi_{t_{0}}}{\partial y_{t_{0}}} \cdot \frac{dy_{t_{0}}}{dt}$, where the terms $\frac{\partial\phi_{t_{0}}}{\partial(x_{t_{0}},y_{t_{0}})}$ are  computed by differentiation of the Nystr\"om Extension formula. With known the value of $\frac{d\phi_{t_{0}}}{dt}$ the value of $\phi_{t_{1}}$ is estimated. By repeating the steps 2-5 integration in the Diffusion Maps coordinates is achieved. }
 
 \label{fig_back_and_forth}
 \end{center}
 \end{figure}

 Given $\frac{d\vect{\phi}_{t_i}}{dt}$, the value $\vect{\phi}_{t_{i+1}}$ can be estimated through any ODE solver. Figure \ref{fig_back_and_forth} illustrates a caricature of the algorithm for a toy example in which the ambient space is $\mathbb{R}^{2}$ and the reduced space $\mathbb{R}$.

\subsection{Grid Tabulation}
\label{sec:Grid_Tabulation}
 The GT algorithm tabulates the reduced vectorfield only for the nodes $\widetilde{\vect{\phi}}_i \in \widetilde{\mathbf{\Phi}}$ in a Cartesian $(N \times N)$ grid. 
 \begin{algorithm}[ht]
 \caption{Grid Tabulation}
 \label{alg:GT}
\DontPrintSemicolon
  \SetKwFunction{FMain}{($\widetilde{\mathbf{\Phi}}$,N)}
  \SetKwFunction{FSum}{Interpolation ($\vect{\phi}_{t_i}$)}

  \SetKwProg{Fn}{Input}{:}{\KwRet}
  \Fn{\FMain}{
  
  1. Lift by using the extension scheme from Algorithm \ref{alg:DDMaps} for each $\widetilde{\vect{\phi}}_i \in \widetilde{\Phi}$. \\
  2. Compute $\frac{d\widetilde{\vect{x}}_i}{dt} = \vect{f}(\widetilde{\vect{x}}_i)$ from the full Equations.\\
  3. For each $\widetilde{\vect{\phi}}_i$ compute $\frac{d\widetilde{\vect{\phi}}_{t_i}}{dt}$, from Equation \eqref{eq:chain rule}.
  
\KwRet $\frac{d\widetilde{\vect{\Phi}}}{dt}$\;
  }\end{algorithm}
Given the tabulated vector field, the integration of the model is performed in the reduced space through a multi-linear interpolation, without the need to map back to the ambient space at each integration step. The use of multi-linear interpolation is not significant: any interpolation scheme could have been used instead. The GT algorithm is similar to the one presented by Chiavazo \textit{et al.} \cite{Chiavazzo_2014}; the main difference in our work is the use of a {\em global interpolation scheme} for the entire manifold during the computation of the first step in Algorithm \ref{alg:GT}. In \cite{Chiavazzo_2014} the authors estimate the reduced derivative
for each node in the grid via the nearest $k$ neighbors, and then lift through Geometric Harmonics {\em only on those $k$ neighbors} (what we here term as Latent Harmonics). This then involves training a regression scheme for every point in the grid separately and not all at once,  as in our case. 

\subsection{Tabulation with Latent Harmonics Interpolation (TaLHI)}
\label{sec:DDMAps Tabulation}
Our last proposed algorithm, TaLHI, learns the reduced vector field as a function of the reduced Diffusion Maps coordinates by using the Latent Harmonics regression scheme. For TaLHI the estimation of the $\frac{d\mathbf{\Phi}}{dt}$ is performed as follows, \hbox {Algorithm \ref{alg:TaLHI}}:

\begin{algorithm}[h]
\caption{TaLHI}
\label{alg:TaLHI}
  \SetKwFunction{FMain}{Input ($\mathbf{X}$)}
  \SetKwProg{Fn}{function}{:}{}
  \Fn{\FMain}{

    1. Compute $\frac{d\vect{x}_i}{dt} = f(\vect{x}_i)$  for each $\vect{x}_i \in \mathbf{X}$. \\
    2. Compute $\frac{d\vect{\phi}_i}{dt}$ by using Equation \eqref{eq:chain rule}. 
    
 \KwRet $\frac{d\mathbf{\Phi}}{dt}$}
\end{algorithm}

The main difference of TaLHI with the other three discussed algorithms is that it doesn't utilize a \textit{lifting} step for the estimation of the reduced derivative since, for each $\vect{\phi}_i \in \mathbf{\Phi}$ we already have the corresponding data point $\vect{x}_i$ in the ambient space. Latent Harmonics regression is only used to construct a mapping from the Diffusion Maps coordinates to their reduced vector field $\frac{d\mathbf{\Phi}}{dt}$.

\section{Results}
We test our suggested algorithms described in Sections \ref{sec:Back-Forth}-\ref{sec:DDMAps Tabulation} through two examples. The first example in Section \ref{sec:Chafee_Infante} is the Chafee-Infante reaction diffusion PDE. The second example, Section \ref{sec:Combustion}, comes from Chemical Kinetics and more precisely from the combustion of hydrogen \cite{Chiavazzo_2014}.

\subsection{The Chafee-Infante PDE}
\label{sec:Chafee_Infante}
To illustrate how reduced dynamical models can be obtained for dissipative PDEs, we utilize the Chafee-Infante reaction-diffusion equation
\begin{equation}
\label{eq:Chafee_Infantee_PDE}
u_t = u - u^3 + \nu{u_{xx}}
\end{equation}
with $\nu$ = 0.16, and boundary conditions \hbox{$u(0,t) = u(\pi,t) = 0$}. This PDE for $\nu =0.16$ it is known that has a two-dimensional inertial manifold \cite{sonday-2011,gear-2011,JollyInertialManifolds}. Diffusion Maps can reveal this two-dimensional manifold embedded in a higher dimensional space given data \textbf{X} sampled from simulations of this PDE \cite{sonday-2011,gear-2011}.
More precisely, to sample data, $\textbf{X}$, firstly we approximate $u$ with a discrete sine transformation:
\begin{equation}
    u(x,t) \approx \sum_{k = 1}^{10} \alpha_k(t)sin(kx)
\end{equation}

 The choice of 10 terms is not significant, results are similar for other levels of accuracy in the approximation \cite{gear-2011}. The Galerkin projection onto those 10 Fourier modes \{sin(x), $\dots$, sin(10x)\}, result in a system of equations of the form:
\begin{equation}
    \label{eq:ode_scheme}
    \frac{d\vect{\alpha}}{dt} = \vect{f}(\vect{\alpha})
\end{equation}
where $\vect{\alpha} \in \mathbb{R}^{10}$.
To sample points on the inertial manifold we integrate this system of ODEs, Equation \ref{eq:ode_scheme}, for different initial conditions $\vect{\alpha}_{t_0}$ and keep only the long-term dynamics. Since we would like to ensure that our sample is as uniform as possible in the ambient space we perform a subsampling where data within a threshold distance $d$ are removed. Thus a uniform data set $\mathbf{X}$ with $\sim 2800$ data points is obtained.

\begin{figure}[ht]
\begin{center}
\centerline{\includegraphics[scale=0.35]{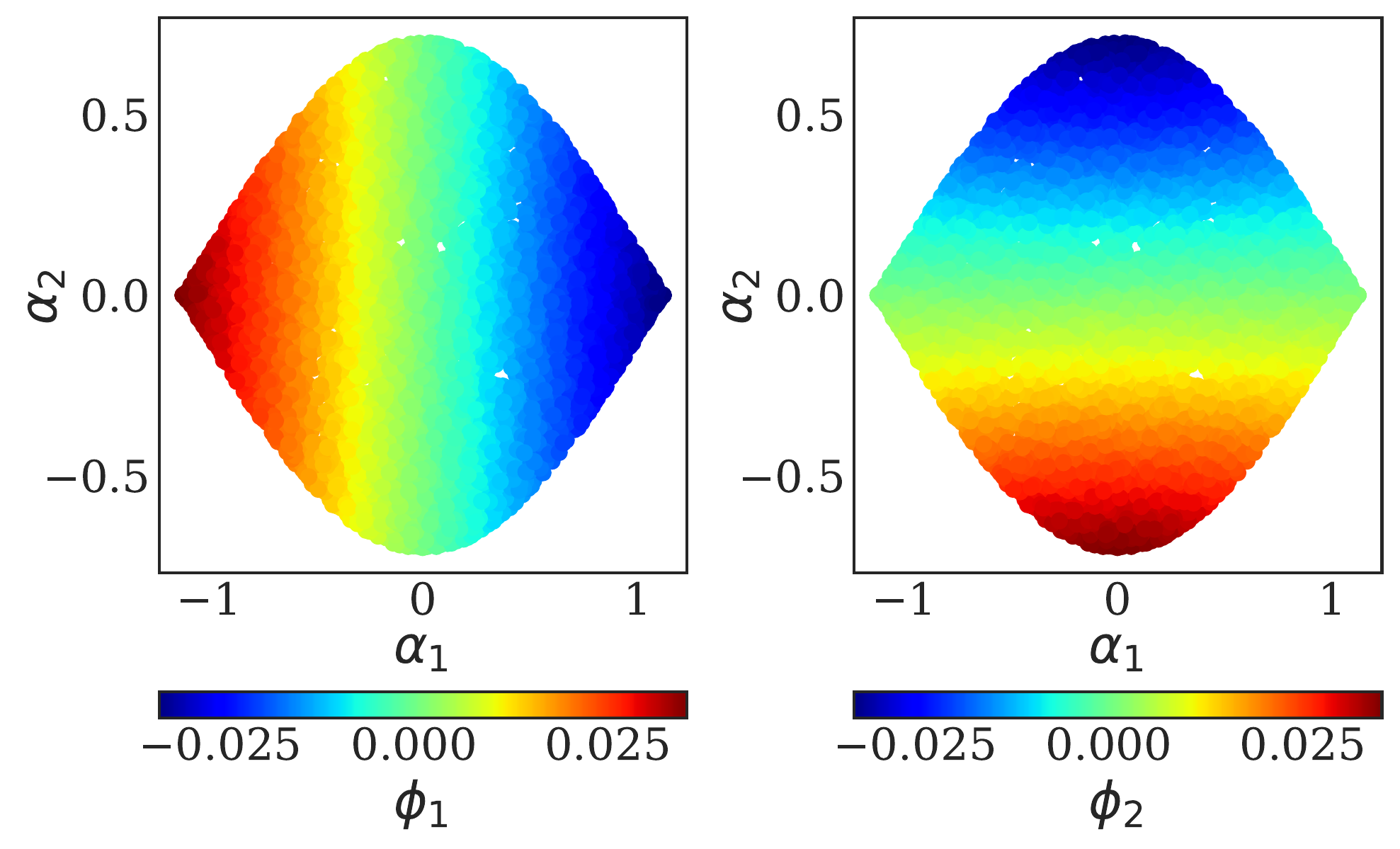}}
\caption{The first two Fourier modes $\alpha_1$ and $\alpha_2$ are colored with the two independent Diffusion Maps coordinates $\phi_1$ (left) and $\phi_2$ (right). The Fourier mode $\alpha_1$ is one-to-one with $\phi_1$ and $\alpha_2$ is one-to-one with $\phi_2$.}
\label{fig:Fourier modes colored}
\end{center}
\end{figure}

Algorithm \ref{alg:DMaps} computed on \textbf{X} reveals a 2-D manifold embedded in $\mathbb{R}^{10}$. The Diffusion Maps independent coordinates, $\phi_1$, $\phi_2$, that fully parametrize this 2-D manifold are also one-to-one with the first two Fourier modes $\alpha_1$,$\alpha_2$ as it can be seen in Figure \ref{fig:Fourier modes colored}. The selected hyperparameters used for Diffusion Maps are reported in Section \ref{sec:DMaps_specifics} of the Appendix. For the Diffusion Maps computations the \textit{datafold} package was used \cite{Lehmberg2020}.

\begin{figure}
     \centering
     \begin{subfigure}[b]{0.7\textwidth}
         \centering
         \includegraphics[width=\textwidth]{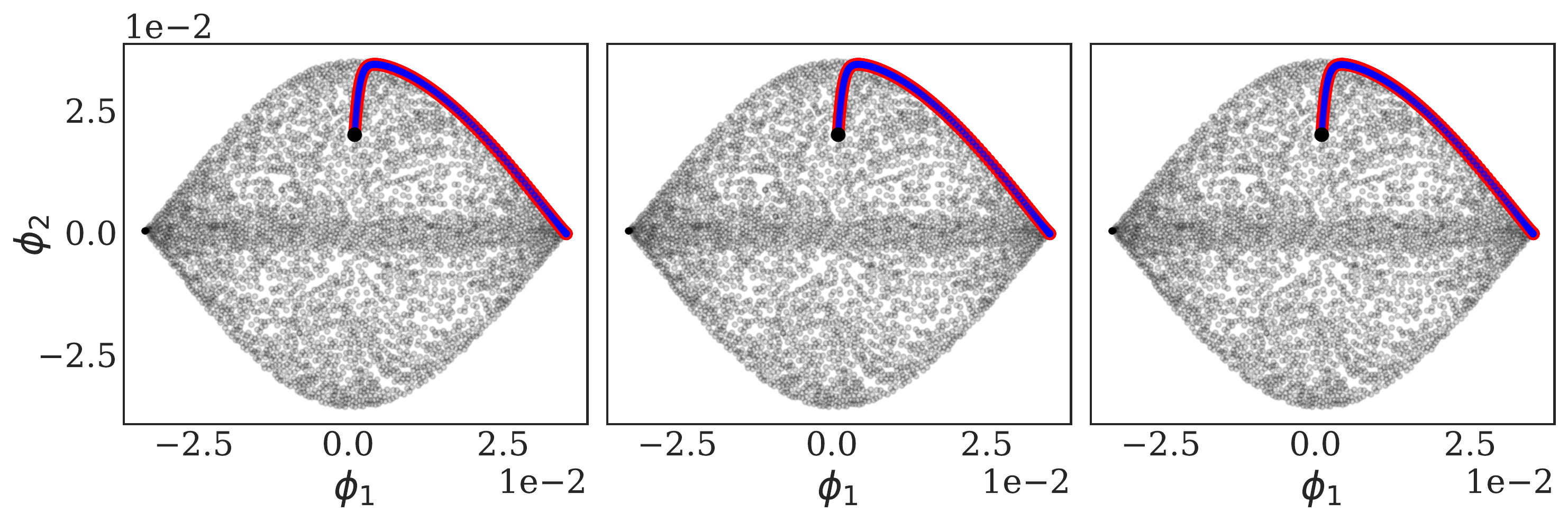}
         \caption{}
\label{fig:LowDimensionTrajectory_integration}       
     \end{subfigure}
     \hfill
     \begin{subfigure}[b]{0.22\textwidth}
         \centering
         \includegraphics[width=\textwidth]{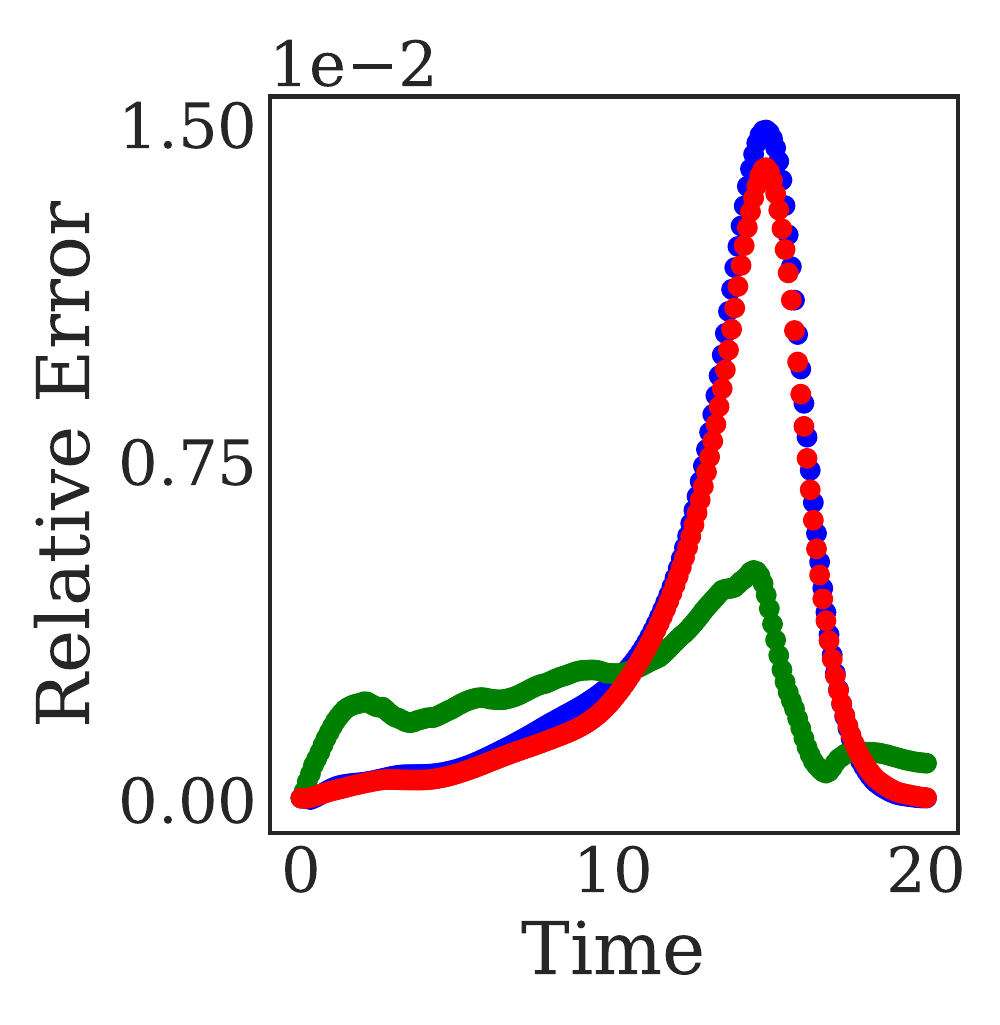}
         \caption{}
\label{fig:LowDimensionTrajectory_error}       
     \end{subfigure}
    \caption{(a) A comparison between a trajectory computed with the three data-driven frameworks TaLHI, BF, and GT, respectively is shown (blue color) against with a restricted (with Nystr\"om Extension) trajectory obtained from the spectral ODEs (red color) for the Chafee-Infante example. (b) The relative error between the data-driven trajectories and the reduced trajectory from the unreduced equations.}
\label{fig:LowDimensionTrajectory}       
\end{figure}

In Figure \ref{fig:LowDimensionTrajectory_integration},  we illustrate a qualitative comparison between the proposed data-driven integration approaches and the restricted (with Nystr{\"o}m extension) solution computed from integrating the original spectral ODEs for a single trajectory. In addition, in Figure \ref{fig:LowDimensionTrajectory_error} the relative error between the restricted trajectory and the three data-driven trajectories is shown in time.

\begin{figure}
     \centering
     \begin{subfigure}[b]{0.6\textwidth}
         \centering
         \includegraphics[width=\textwidth]{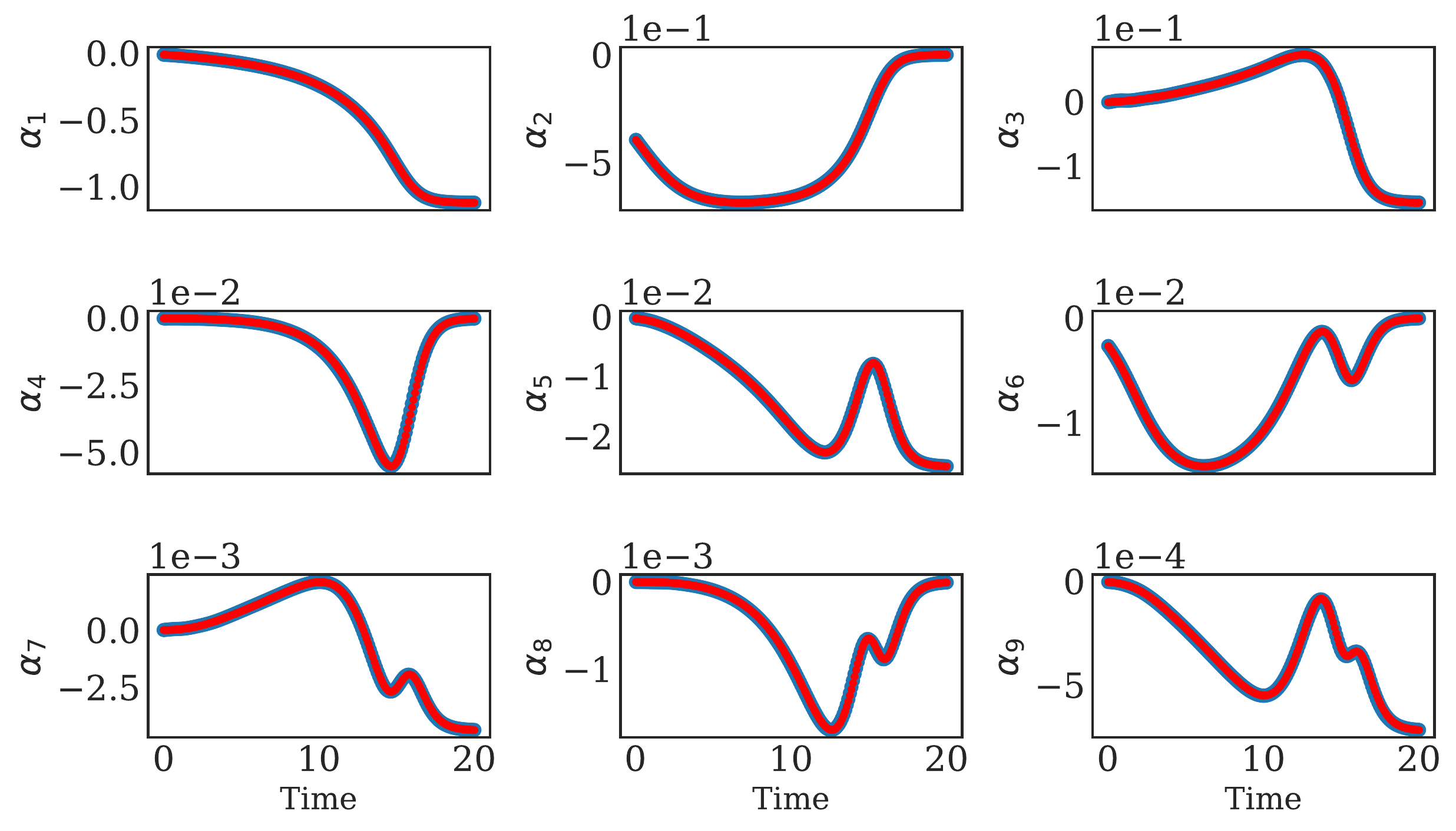}
         \caption{}
\label{fig:High Dimensonal Trajectory Integration}    
     \end{subfigure}
     \hfill
     \begin{subfigure}[b]{0.25\textwidth}
         \centering
         \includegraphics[width=\textwidth]{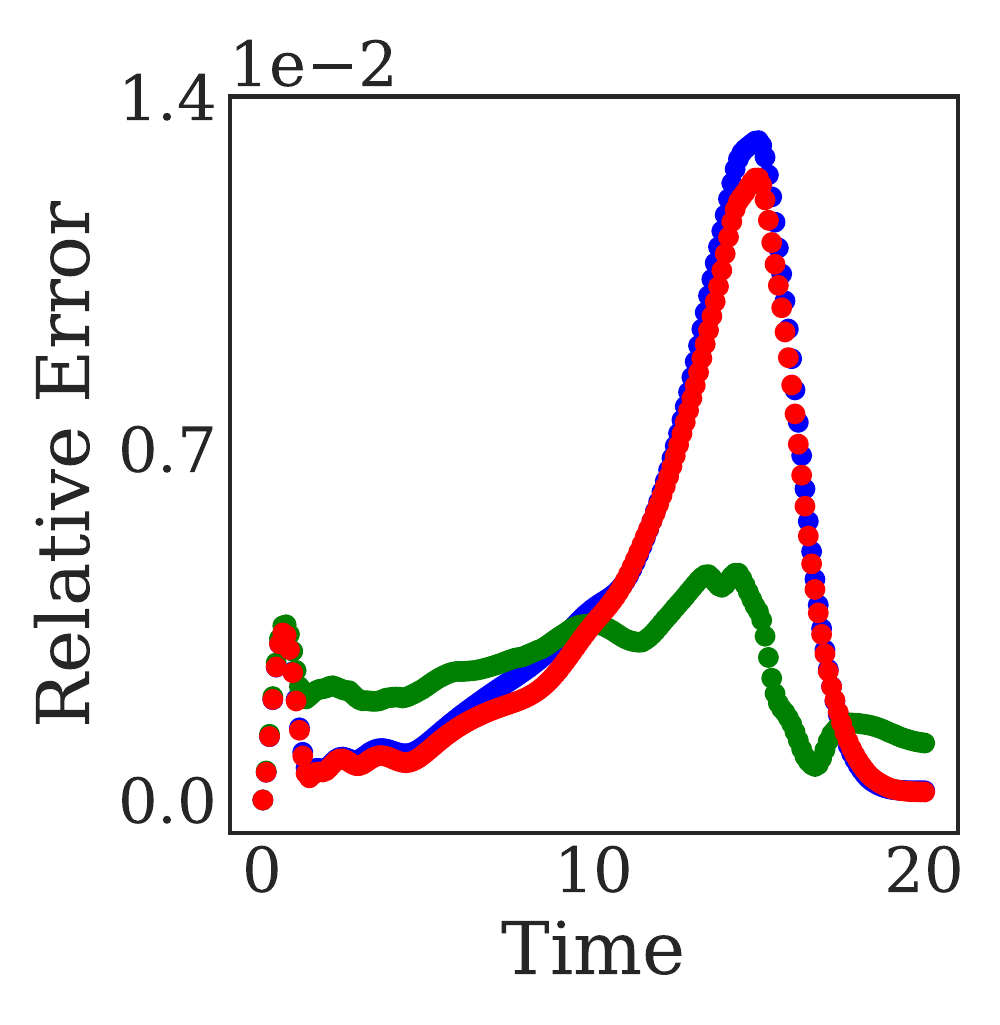}
         \caption{}
\label{fig:High Dimensonal Trajectory Error}    
     \end{subfigure}
\caption{(a) A comparison between the lifted with Latent Harmonics trajectory computed
from the data-driven integration scheme TaLHI (blue color) and the  trajectory computed from the full, unreduced equations (red color) is shown for the first nine Fourier Modes. (b) The relative errors in time between the data-driven trajectories and the trajectory from the full, unreduced equations,  BF (blue), GT (green), TaLHI (red).}
\label{fig:High Dimensonal Trajectory}    
\end{figure}

A comparison between the data-driven trajectories in the ambient space is also possible. Figure \ref{fig:High Dimensonal Trajectory Integration} shows the comparison between the lifted trajectory computed with TaLHI algorithm and the first 9 components for a trajectory coming from the full, unreduced equations. Qualitative comparisons in the ambient space for the BF and GT algorithms are shown in \hbox{ Section \ref{sec:Additional_Results}} of the Appendix.  In Figure \ref{fig:High Dimensonal Trajectory Error} also the relative error in time between the trajectories computed with the data-driven algorithms and the trajectory from the full, unreduced equations is  shown in time.

\begin{figure}[ht]
\begin{center} 
\includegraphics[scale =0.34]{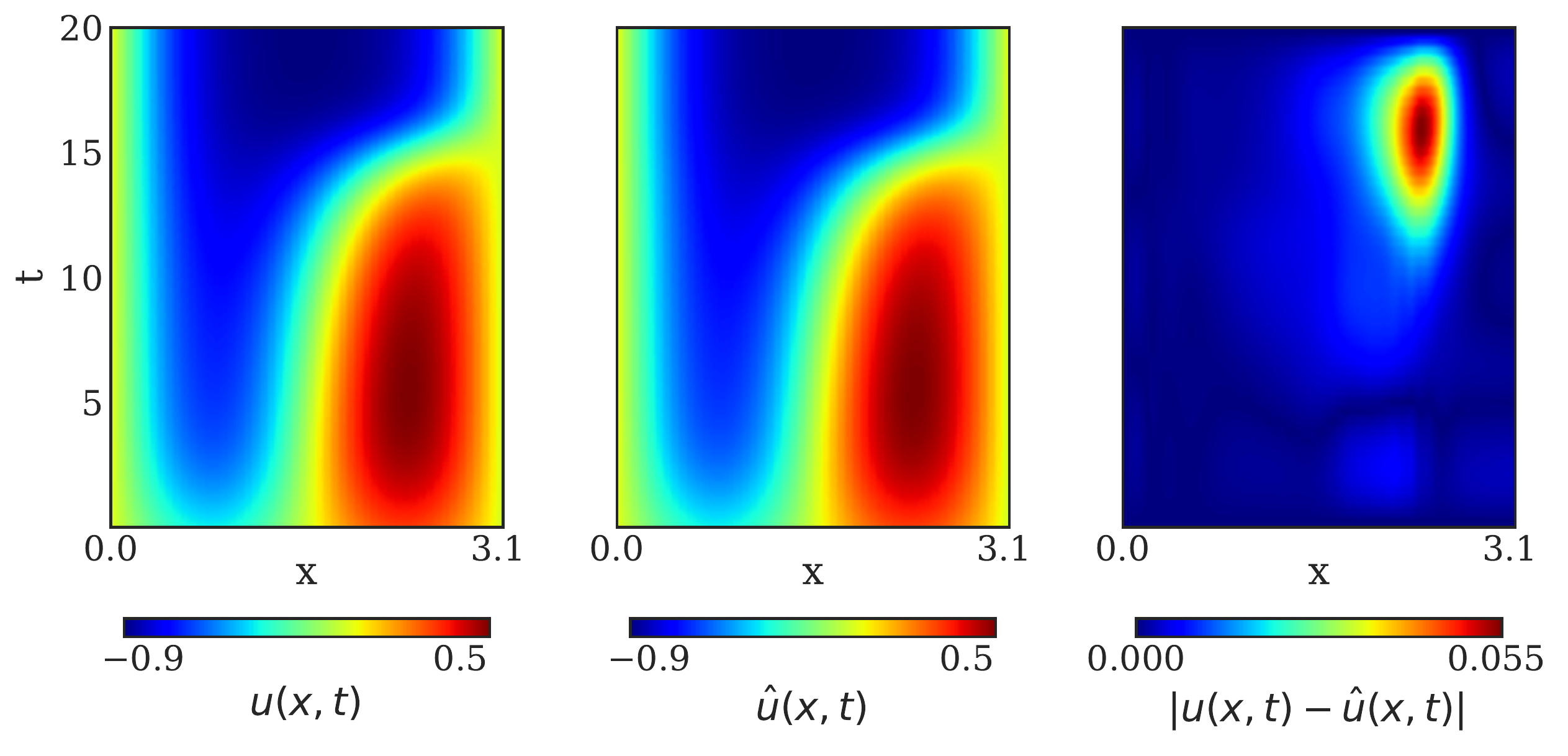}
\caption{A trajectory $u(x,t)$ of the Chafee Infante PDE, Equation  \eqref{eq:Chafee_Infantee_PDE}, integrated with Finite Differences (left) compared to a lifted \& reconstructed trajectory $\hat{u}(x,t)$ obtained with the data-driven scheme TaLHI. The absolute error of $u(x,t)$ and the data-driven trajectory $\hat{u}(x,t)$ is shown in the right.}
\label{fig:Full_PDE}
\end{center}
\end{figure}

The reconstruction from the latent coordinates back to the original $u(x,t)$ space of the PDE in Equation \eqref{eq:Chafee_Infantee_PDE} is also possible through our scheme. In Figure \ref{fig:Full_PDE} the comparison between a trajectory of the Chafee-Infante PDE is contrasted to a lifted and reconstructed trajectory with the same initial condition, integrated with the data-driven scheme TaLHI in the latent coordinates.

\subsection{Chemical Kinetics}
\label{sec:Combustion}
The second application with which we want to test our algorithms is a system of stiff ODES. This example involves constructing a reduced kinetic model for a combustion example. This system of equations describes the combustion of hydrogen and air at stoichiometric proportions under fixed total enthalpy (H = 300 $\frac{k\text{J}}{\text{kg}}$) 
and pressure (P = 1 bar). For this dynamical system the long term dynamics live in a two-dimensional manifold \cite{Chiavazzo_2014}. We used the detailed mechanism of Li et al. \cite{Limechanism}, in which nine chemical species (\ce{H2}, \ce{N2}, \ce{H}, \ce{O}, \ce{OH}, \ce{O2}, \ce{H2O}, \ce{HO2}, \ce{H2O2}) and three elements (\ce{H}, \ce{O}, \ce{N}) are participating, with the concentration of \ce{N2} to remain constant during the reaction. The selection of this second example on Chemical Kinetics and our Grid Tabulation (GT) discussed in Section \ref{sec:Grid_Tabulation} approach aims to give a comparison between our proposed scheme and the work of Chiavazzo et al \cite{Chiavazzo_2014}. We have used the same model and also the same sampling strategy trying to mimic the work of Chiavazzo et al \cite{Chiavazzo_2014} to collect a data set $\mathbf{X}$ of $\sim$ 3800 data points.
 
\begin{figure}[ht]
\begin{center} 
\includegraphics[scale=0.4]{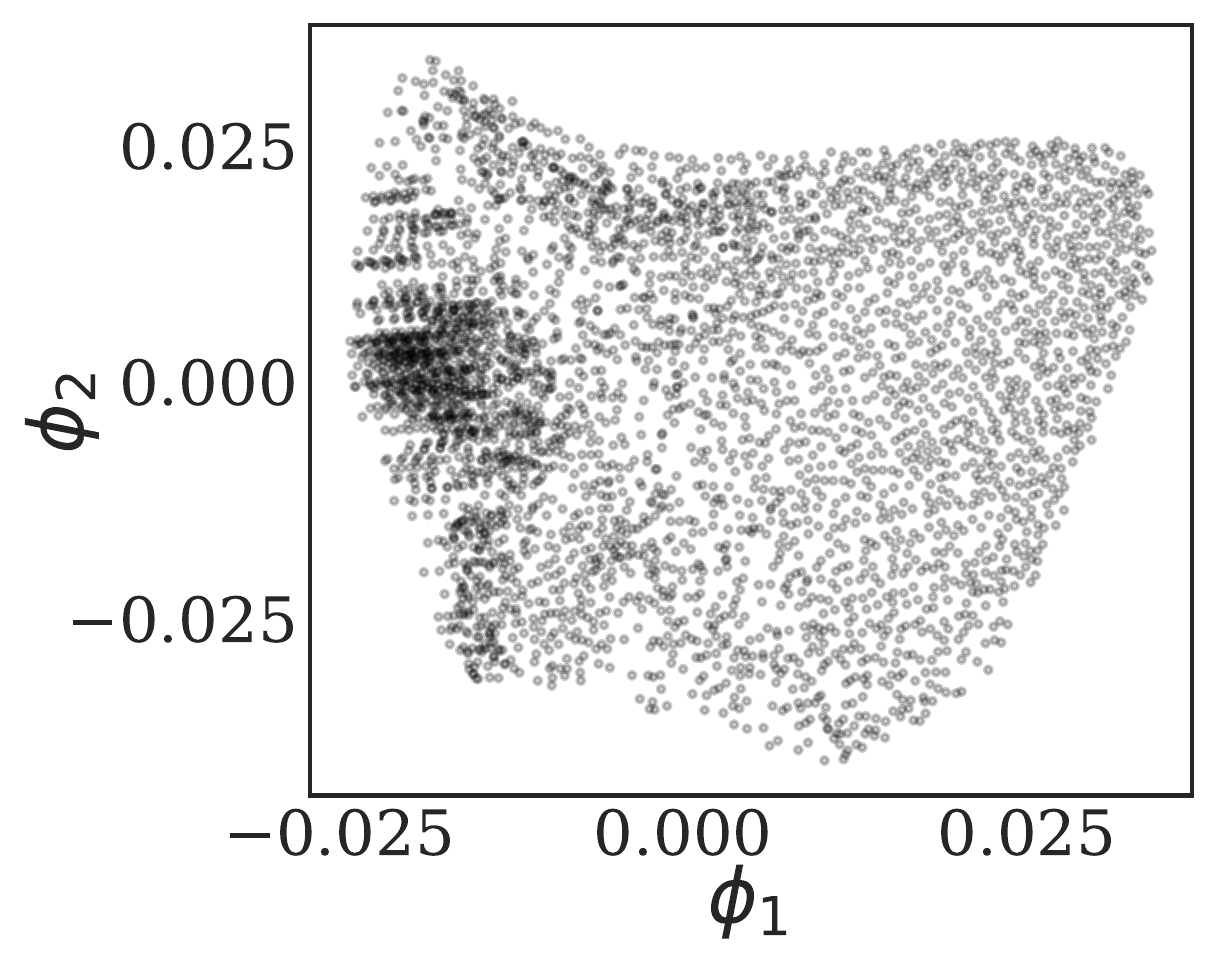}
\caption{The two dimensional manifold in terms of the independent eigenvectors $\phi_1$ and $\phi_2$ discovered on Chemical Kinetics example.}
\label{fig:Combustion_DMAPs}
\end{center}
\end{figure}

 Algorithm \ref{alg:DMaps} applied on our collected data discovers a two-dimensional manifold embedded in $\mathbb{R}^9$, Figure \ref{fig:Combustion_DMAPs}. On those latent coordinates we test our schemes for constructing reduced dynamical models. It is worth mentioning that this system of equations is stiff because of the widely disparate time scales, and also because the concentrations of some chemical species change by multiple orders of magnitude. This makes the global interpolation between the Diffusion Maps space and the ambient space numerically challenging. It is worth reminding the reader that the  main difference between our lifting scheme and the lifting used with Latent Harmonics in \cite{Chiavazzo_2014} is that in our approach we build our extension scheme by parametrizing {\em the entire manifold} in contrast to the local approach (nearest neighbors) used in \cite{Chiavazzo_2014}. Therefore Latent Harmonics in our case attempts a more challenging approach, namely global learning and extension of stiffer functions.  In our case to learn the dynamics the \textit{training} discussed in Algorithm \ref{alg:DDMaps} needs to be performed only once given all the available data points sampled on the manifold; then any point in the \textit{neighborhood} can be \textit{lifted}. In the work of Chiavazzo et al. \cite{Chiavazzo_2014}  the \textit{training} of the for \textit{lifting} any point of interest needs to be performed separately for each point. A comparison between the computational time needed to construct a $60 \times 60$ grid with our GT approach (12.50 mins) and the approach discussed in  Chiavazzo et al (45.10 mins) was performed (both experiments conducted on a Intel(R) Xeon(R) Silver 4112 CPU @ 2.60GHz Processor) and shows a speed up of for our tabulation scheme with GT.

\begin{figure}
     \centering
     \begin{subfigure}[b]{0.7\textwidth}
         \centering
         \includegraphics[width=\textwidth]{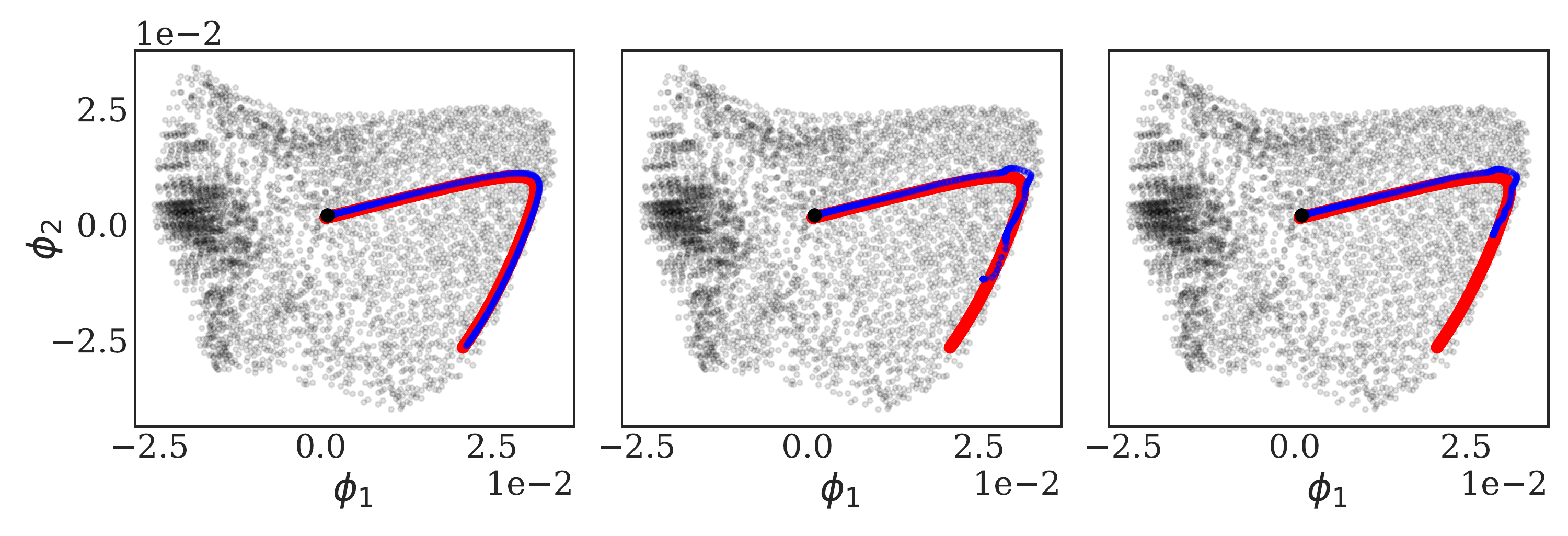}
         \caption{}
         \label{fig:Reduced_Integration_Combustion}
     \end{subfigure}
     \hfill
     \begin{subfigure}[b]{0.23\textwidth}
         \centering
         \includegraphics[width=\textwidth]{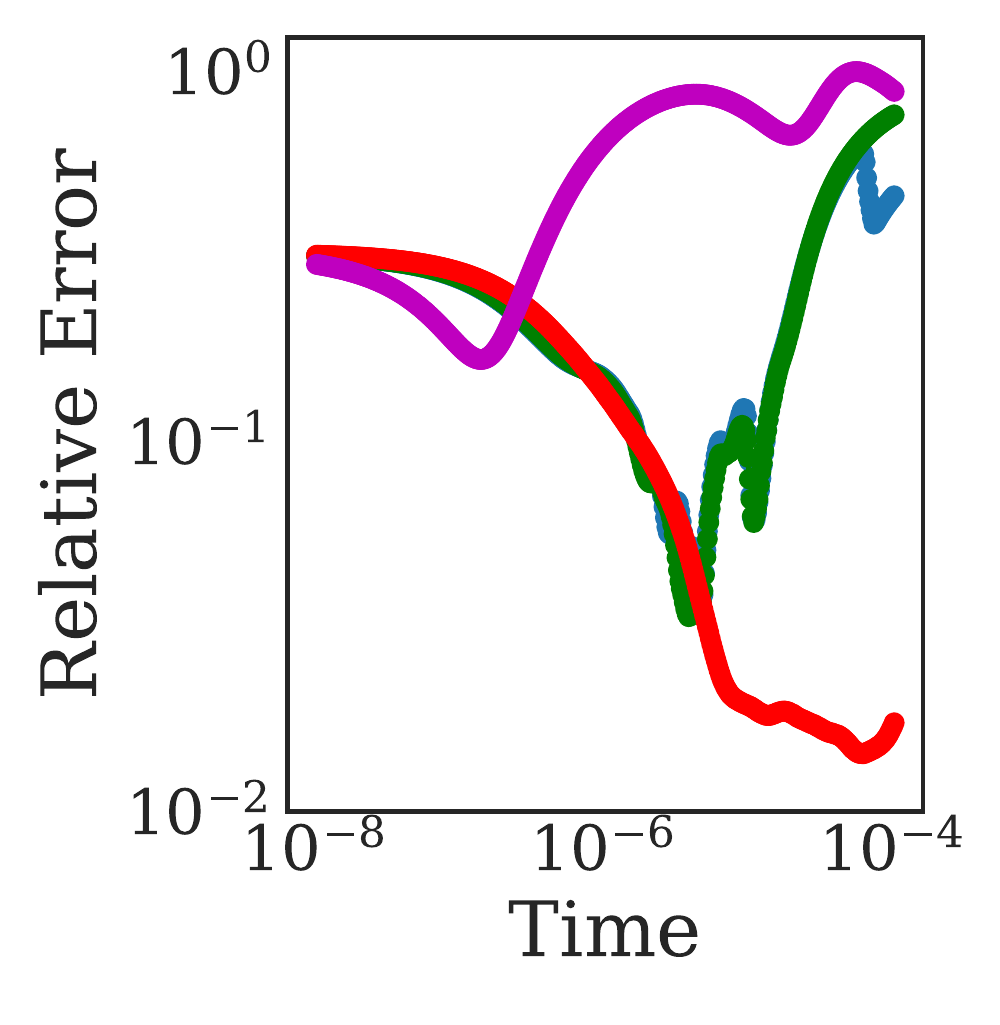}
         \caption{}
    \label{fig:Reduced_Error_Combustion}
     \end{subfigure}
\caption{(a) A comparison between a trajectory computed from the three data-driven frameworks TaLHI, BF, and GT, respectively is shown (blue color) and a restricted (with Nystr\"om Extension) trajectory obtained from the spectral ODEs (red color) for the example from Chemical Kinetics.(b) The relative error in time of the four data driven approaches for each point along the simulated trajectory, BF (blue), GT (green), TaLHI (red), Chiavazzo et al \cite{Chiavazzo_2014} (mangenta), each compared to the restricted with Nystr\"om Extension trajectory of the full dynamics.}
\label{fig:Combustion_DMAPs_Low_Dimensional}
\end{figure}

\begin{figure}
     \centering
     \begin{subfigure}[b]{0.6\textwidth}
         \centering
         \includegraphics[width=\textwidth]{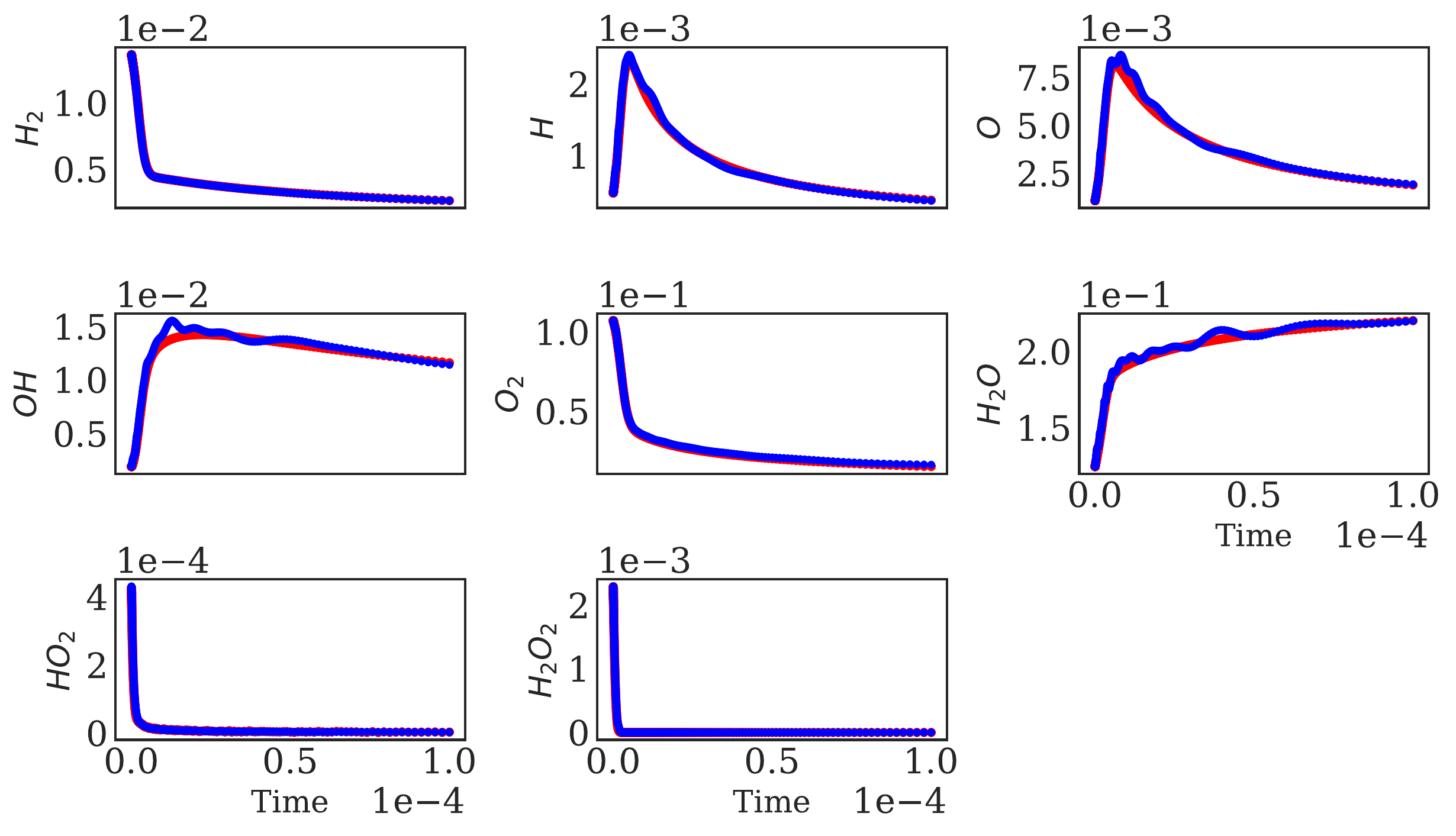}
         \caption{}
\label{fig:CO_DMAPs_High_Dimensional_TaLHI_integration}
     \end{subfigure}
     \hfill
     \begin{subfigure}[b]{0.23\textwidth}
         \centering
         \includegraphics[width=\textwidth]{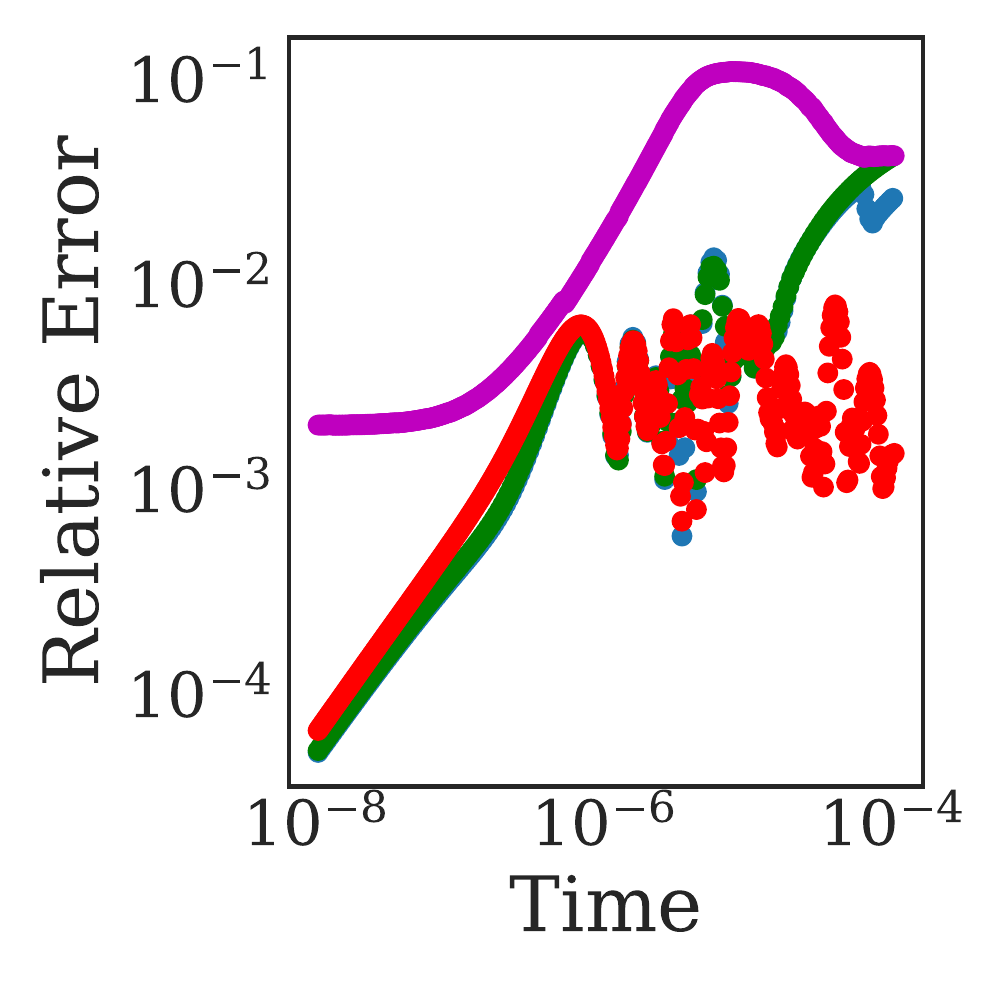}
         \caption{}
\label{fig:CO_DMAPs_High_Dimensional_TaLHI_error}
     \end{subfigure}
\caption{(a) Comparison  between the lifted with Latent Harmonics trajectory computed with the data-driven  integration  scheme Back and Forth (blue trajectory) and  the trajectory computed from the full, unreduced equations of the kinetics (red trajectory). (b) The relative error in the ambient space between the lifted data driven schemes and a trajectory of the full dynamics, BF (blue), GT (green), TaLHI (red), Chiavazzo et al \cite{Chiavazzo_2014} (mangenta). The lifting in our proposed data-driven schemes was performed with Algorithm \ref{alg:DDMaps} and the one in Chiavazzo et al. with \textit{local} Geometric Harmonics.}
\label{fig:CO_DMAPs_High_Dimensional_TaLHI}
\end{figure}

 The goal for this example remains the same as before, to construct a data-driven integration scheme in the latent coordinates. Figure \ref{fig:Reduced_Integration_Combustion}
illustrates a qualitative comparison between the restricted with Nystr\"om full, unreduced equations and the three proposed data-driven approaches. In Figure \ref{fig:Reduced_Error_Combustion} the relative error between the trajectory of the restricted equations and the data-driven trajectories is shown. We computed also the trajectory with the approach discussed in Chiavazzo et al. \cite{Chiavazzo_2014} to give a more quantitative comparison between our suggested approach an the one in \cite{Chiavazzo_2014}.Figure \ref{fig:CO_DMAPs_High_Dimensional_TaLHI_integration} shows the comparison between the lifted data-driven trajectory integrated with TaLHI and a trajectory of the full system of equations. Qualitative comparisons between the trajectory computed from the full model and trajectories computed with the data-driven integration schemes BF and GT in the ambient space are given in Section \ref{sec:Additional_Results} of the Appendix. The relative error between the trajectory of the full equations and lifted trajectories from our suggested data-driven schemes including the one from Chiavazzo et al \cite{Chiavazzo_2014} for each point along the simulated trajectories are shown in Figure \ref{fig:CO_DMAPs_High_Dimensional_TaLHI_error}.

\section{Discussion-Conclusions}
\label{sec:Conclusions}
We presented three different approaches that allow us to construct reduced data-driven models in the Diffusion Maps coordinates. The key elements for constructing reduced data-driven models thought the Double Diffusion Maps scheme were described in Algorithms \ref{alg:DMaps} and \ref{alg:DDMaps} along with the Nystr\"om Extension, discussed in detail in Section \ref{sec:Nystrom Extension} of the Appendix. We tested our algorithms in two different examples in which the long term dynamics lie on a low-dimensional, slow-inertial manifold. To test our Double Diffusion Maps scheme we discover a set of latent coordinates and test our suggested schemes for the construction of reduced dynamical models. We compared the data-driven trajectories to the dynamics of the closed form expressions either in the latent coordinates or in the ambient space, In the ambient space we used the Latent Harmonics (Algorithm \ref{alg:DDMaps}) to lift our reduced trajectories in the ambient space. The Comparison in the reduced coordinates was achieved by restricting the full trajectories computed from the equations in the reduced space through Nystr\"om Extension.

The proposed approaches named, BF, GT and, TaLHI have their strengths and weaknesses. The tabulation schemes GT and TaLHI are computationally expensive and need to allocate memory during the tabulation process. However, the tabulation needs to be performed only once.

Based on our examples TaLHI seems to be more accurate compared to the other two approaches. This can be perhaps attributed to the fact that this method does not acquire lifting in order to learn the reduced dynamics. 

The BF approach avoids the expensive tabulation step. Integrating in the reduced coordinates without the need to tabulate in advance the vector field would be even more beneficial in cases the reduced space coordinates are higher than two. In addition, the BF approach was constructed under the assumption that by decreasing the dimensions of a dynamical system its stiffness also decreases. This seems to be valid in our examples, and so  larger time steps can be used in the reduced space compared to the ambient space. 

It seems that whenever the trajectory visits the boundary of a manifold it becomes more prone to numerical errors. Additional sampling in the boundary of the manifold will, we expect, increase the interpolation accuracy of our scheme and thus give more accurate dynamic models \cite{BoundaryAnastasia} . 

\begin{figure}[h]
    \begin{center}
    \includegraphics[scale=0.45]{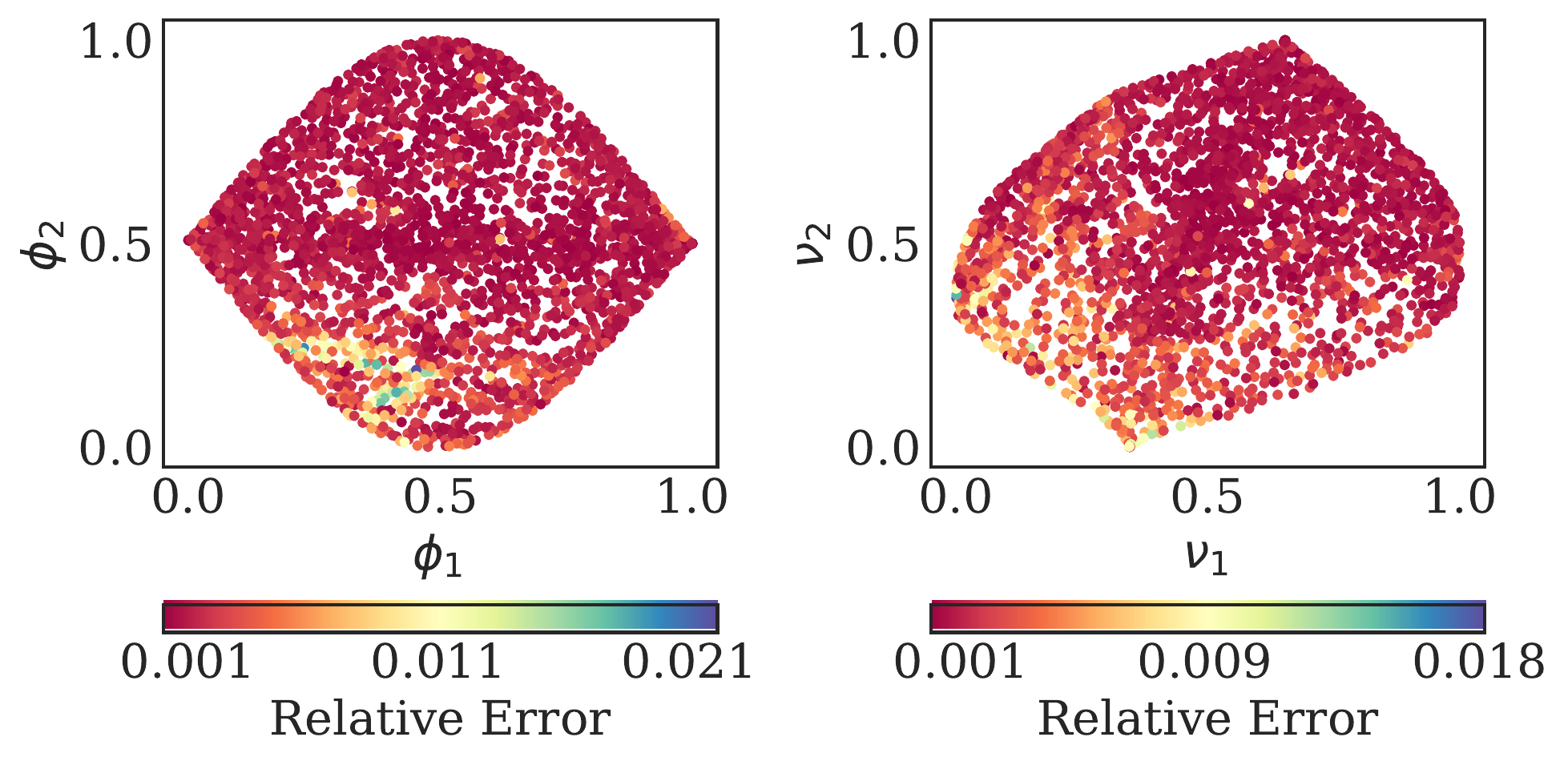}
    \caption{The relative error $\frac{|\vect{\phi}_i - \hat{\vect{\phi}}_i|}{\vect{\phi}_i}$ of the forward mapping, $f: N \to \Phi$ is shown on the left as a function on the original values of the Diffusion Maps coordinates. The relative error $\frac{|\vect{\nu}_i - \hat{\vect{\nu}}_i|}{\vect{\nu}_i}$ of the backward mapping $f^{-1}: \Phi \to N $ is shown on the right. In both cases the coordinates are rescaled in the range 0 to 1.}
    \label{fig:One_to_One}
    \end{center}
    \end{figure}

In our current work we chose, for discovering the latent coordinates, the manifold learning method Diffusion Maps. Of course, other non-linear dimensionality reduction methods could have been used alternatively. To justify that we trained an autoencoder neural network with two bottle neck variables for the Chafee-Infantee example discussed in Section \ref{sec:Chafee_Infante}. The two latent coordinates computed from the Autoencoder ($\nu_1, \nu_2$) are one-to-one with the Diffusion Maps variables. The one-to-one relationship between the two sets of variables was confirmed by learning a mapping $f:N \to \Phi$ with Geometric Harmonics and the inverse mapping $f^{-1}:\Phi \to N$ with Latent Harmonics. In Figure \ref{fig:One_to_One} we plot the relative error computed for each test set point as function of the (rescaled to the range 0 to 1) Diffusion Maps Coordinates and Autoencoders Coordinates respectively.
We believe that the natural use of the intrinsic geometry dictated by the data, discovered through an eigenproblem, is 
``more direct" than an autoencoder-based latent space, even if both lead to comparable reduction.
Indeed, the intrinsic-geometry based approach will lead always to the same reduced space parametrization for the same data; an autencoder based one will/may lead to a different (equivalent) representation for different researchers/different optimization runs, unless a systematic regularization is used.

\begin{figure}[h]
 \begin{center}
\includegraphics[scale=0.45]{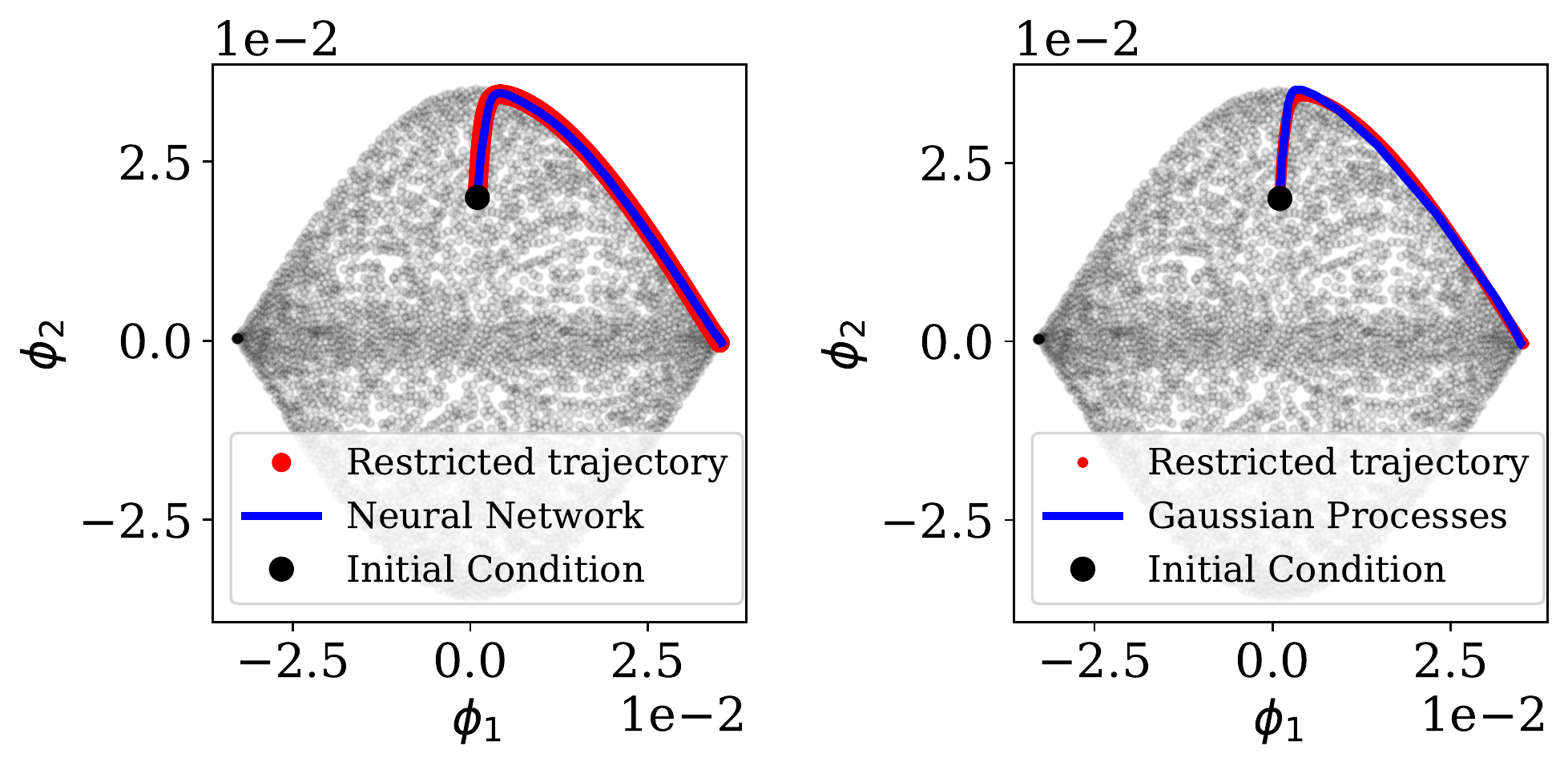}
\caption{data-driven Integration with a Neural Network (on the left) and Gaussian Processes (on the right).}
\label{fig:GPr_NN_trajectories}
\end{center}
\end{figure}

In our work since the reduced models are constructed in the Diffusion Maps coordinates it made sense to use a regression scheme based on its extension; however different regression algorithms could have also been used instead. In Figure \ref{fig:GPr_NN_trajectories} we show the ability to learn the reduced dynamics in the Diffusion Maps coordiantes and integrate with Gaussian Proccesses and also with a Neural Network.

We conclude, by mentioning that the focus of this paper was on the construction of reduced models for dynamical systems whose long term behavior live in a lower dimensional manifold. We illustrated three possible algorithms with which a reduced data-assisted model can be constructed. Those suggested algorithms could be used to any dynamical systems of ordinary or partial differential equation with reduced long term behavior. Those data-driven models can be used as a more efficient, compressed simulator simulator in cases that using the full model repeatedly is computationally inefficient (e.g. control processing). In our algorithms, we assumed that sampling the whole portion of interest of the manifold is necessary; however this might not be possible or computationally efficient. In problems in science and engineering data might be given only in a small neighborhood of their entire manifold whose dimensionality also might not be known in advance. In that direction further research might needed in order to utilize \textit{effective} sampling algorithms combined with algorithms that aim to learn operators on the data \cite{chiavazzo2016imapd}. 

In addition, the latent coordinates in our case were computed without taking into consideration physical constrains that may exist in their original model (e.g. mass conservation during chemical reactions). Infusing such constraints in physics-informed models  \cite{raissi2019physics} constitutes another promising research area.

\paragraph{Funding:} This work was partially supported by the US AFOSR and the US Department of Energy.

\paragraph{Acknowledgements:} N.E with the help of F.D, performed the Diffusion Maps computations and developed the code for the three suggested schemes for the construction of reduced data-driven models. N.E, F.D, and I.G.K wrote the paper. E.C, provided data and models. All authors designed and performed research. All authors contributed in manuscript revision.

\bibliographystyle{plain}
\bibliography{lit}

\begin{thebibliography}{10}

\bibitem{benner2015survey}
Peter Benner, Serkan Gugercin, and Karen Willcox.
\newblock A survey of projection-based model reduction methods for parametric
  dynamical systems.
\newblock {\em SIAM review}, 57(4):483--531, 2015.

\bibitem{TimeScaleSeparation}
Tyrus Berry, John Cressman, Z.~Gregurić-Ferenček, and T.~Sauer.
\newblock Time-scale separation from diffusion-mapped delay coordinates.
\newblock {\em SIAM Journal on Applied Dynamical Systems [electronic only]},
  12, 01 2013.

\bibitem{champion2019data}
Kathleen Champion, Bethany Lusch, J~Nathan Kutz, and Steven~L Brunton.
\newblock Data-driven discovery of coordinates and governing equations.
\newblock {\em Proceedings of the National Academy of Sciences},
  116(45):22445--22451, 2019.

\bibitem{chen2019selecting}
Yu-Chia Chen and Marina Meila.
\newblock Selecting the independent coordinates of manifolds with large aspect
  ratios.
\newblock {\em Advances in Neural Information Processing Systems}, 32, 2019.

\bibitem{chiavazzo2016imapd}
Eliodoro Chiavazzo, Roberto Covino, Ronald~R. Coifman, C.~William Gear,
  Anastasia~S. Georgiou, Gerhard Hummer, and Ioannis~G. Kevrekidis.
\newblock Intrinsic map dynamics exploration for uncharted effective
  free-energy landscapes.
\newblock {\em Proceedings of the National Academy of Sciences},
  114(28):E5494--E5503, 2017.

\bibitem{Chiavazzo_2014}
Eliodoro Chiavazzo, Charles Gear, Carmeline Dsilva, Neta Rabin, and Ioannis
  Kevrekidis.
\newblock Reduced models in chemical kinetics via nonlinear data-mining.
\newblock {\em Processes}, 2(1):112–140, Jan 2014.

\bibitem{CoifmanTomographyLaplacian}
R.~R. {Coifman}, Y.~{Shkolnisky}, F.~J. {Sigworth}, and A.~{Singer}.
\newblock Graph laplacian tomography from unknown random projections.
\newblock {\em IEEE Transactions on Image Processing}, 17(10):1891--1899, 2008.

\bibitem{COIFMAN20065}
Ronald~R. Coifman and Stéphane Lafon.
\newblock Diffusion maps.
\newblock {\em Applied and Computational Harmonic Analysis}, 21(1):5 -- 30,
  2006.
\newblock Special Issue: Diffusion Maps and Wavelets.

\bibitem{Geometric_harmonics_paper}
Ronald~R. Coifman and Stéphane Lafon.
\newblock Geometric harmonics: A novel tool for multiscale out-of-sample
  extension of empirical functions.
\newblock {\em Applied and Computational Harmonic Analysis}, 21(1):31 -- 52,
  2006.
\newblock Special Issue: Diffusion Maps and Wavelets.

\bibitem{KSE_reaction_diff}
Peter Constantin, Ciprian Foias, Basil Nicolaenko, and Roger Temam.
\newblock {\em Integral manifolds and inertial manifolds for dissipative
  partial differential equations}, volume~70.
\newblock Springer Science \& Business Media, 2012.

\bibitem{deane1991low}
AE~Deane, IG~Kevrekidis, G~Em Karniadakis, and SA~Orszag.
\newblock Low-dimensional models for complex geometry flows: application to
  grooved channels and circular cylinders.
\newblock {\em Physics of Fluids A: Fluid Dynamics}, 3(10):2337--2354, 1991.

\bibitem{Ginzburg-Landau_LadayDoering_1988}
C~R Doering, J~D Gibbon, D~D Holm, and B~Nicolaenko.
\newblock Low-dimensional behaviour in the complex ginzburg-landau equation.
\newblock {\em Nonlinearity}, 1(2):279--309, may 1988.

\bibitem{dsilva2015parsimonious}
Carmeline~J Dsilva, Ronen Talmon, Ronald~R Coifman, and Ioannis~G Kevrekidis.
\newblock Parsimonious representation of nonlinear dynamical systems through
  manifold learning: A chemotaxis case study.
\newblock {\em Applied and Computational Harmonic Analysis}, 44(3):759--773,
  2018.

\bibitem{Inertial_Manifolds_Temam}
Ciprian Foias, George~R Sell, and Roger Temam.
\newblock Inertial manifolds for nonlinear evolutionary equations.
\newblock {\em Journal of Differential Equations}, 73(2):309 -- 353, 1988.

\bibitem{EfficientspatiotemporalNystrom}
C.~{Fowlkes}, S.~{Belongie}, and J.~{Malik}.
\newblock Efficient spatiotemporal grouping using the nystrom method.
\newblock In {\em Proceedings of the 2001 IEEE Computer Society Conference on
  Computer Vision and Pattern Recognition. CVPR 2001}, volume~1, pages I--I,
  2001.

\bibitem{FROUZAKIS200075}
C.E. Frouzakis, Y.G. Kevrekidis, J.~Lee, K.~Boulouchos, and A.A. Alonso.
\newblock Proper orthogonal decomposition of direct numerical simulation data:
  Data reduction and observer construction.
\newblock {\em Proceedings of the Combustion Institute}, 28(1):75--81, 2000.

\bibitem{Gear_Chiavazzo_Parametrizing_boundary_detection}
C.~W. Gear, E.~Chiavazzo, and I.~G. Kevrekidis.
\newblock Manifolds defined by points: Parameterizing and boundary detection
  (extended abstract).
\newblock {\em AIP Conference Proceedings}, 1738(1):020005, 2016.

\bibitem{gear-2011}
C.~W. Gear, I.~G. Kevrekidis, and B.~E. Sonday.
\newblock Slow manifold integration on a diffusion map parameterization.
\newblock {\em AIP Conference Proceedings}, 1389(1):13--16, 2011.

\bibitem{BoundaryAnastasia}
Anastasia Georgiou, Juan Bello-Rivas, Charles Gear, Hau-Tieng Wu, Eliodoro
  Chiavazzo, and Ioannis Kevrekidis.
\newblock An exploration algorithm for stochastic simulators driven by energy
  gradients.
\newblock {\em Entropy}, 19(7):294, Jun 2017.

\bibitem{Ginzburg-Landau_2}
J.~M. Ghidaglia and B.~Her\'{o}n.
\newblock Dimension of the attractors associated to the ginzburg-landau partial
  differential equation.
\newblock {\em Phys. D}, 28(3):282–304, October 1987.

\bibitem{Reaction_Diffusion_Manif}
Michael~S. Jolly.
\newblock Explicit construction of an inertial manifold for a reaction
  diffusion equation.
\newblock {\em Journal of Differential Equations}, 78(2):220 -- 261, 1989.

\bibitem{JollyInertialManifolds}
Michael~S. {Jolly}.
\newblock {Explicit construction of an inertial manifold for a reaction
  diffusion equation}.
\newblock {\em Journal of Differential Equations}, 78(2):220--261, January
  1989.

\bibitem{Jolly_Pde_good_explanations}
M.S. Jolly, I.G. Kevrekidis, and E.S. Titi.
\newblock Approximate inertial manifolds for the kuramoto-sivashinsky equation:
  Analysis and computations.
\newblock {\em Physica D: Nonlinear Phenomena}, 44(1):38 -- 60, 1990.

\bibitem{kevrekidis2004equation}
Ioannis~G Kevrekidis, C~William Gear, and Gerhard Hummer.
\newblock Equation-free: The computer-aided analysis of complex multiscale
  systems.
\newblock {\em AIChE Journal}, 50(7):1346--1355, 2004.

\bibitem{kevrekidis2003equation}
Ioannis~G Kevrekidis, C~William Gear, James~M Hyman, Panagiotis~G Kevrekidid,
  Olof Runborg, Constantinos Theodoropoulos, et~al.
\newblock Equation-free, coarse-grained multiscale computation: Enabling
  microscopic simulators to perform system-level analysis.
\newblock {\em Communications in Mathematical Sciences}, 1(4):715--762, 2003.

\bibitem{kevrekidis2009equation}
Ioannis~G Kevrekidis and Giovanni Samaey.
\newblock Equation-free multiscale computation: Algorithms and applications.
\newblock {\em Annual review of physical chemistry}, 60:321--344, 2009.

\bibitem{koelle2018manifold}
Samson Koelle, Hanyu Zhang, Marina Meila, and Yu-Chia Chen.
\newblock Manifold coordinates with physical meaning.
\newblock {\em arXiv e-prints}, pages arXiv--1811, 2018.

\bibitem{krischer1993model}
K~Krischer, R~Rico-Mart{\'\i}nez, IG~Kevrekidis, HH~Rotermund, G~Ertl, and
  JL~Hudson.
\newblock Model identification of a spatiotemporally varying catalytic
  reaction.
\newblock {\em AIChE Journal}, 39(1):89--98, 1993.

\bibitem{Kuhn_Data_Driven_Modeling_Scientific_Computations}
J.~Nathan Kutz.
\newblock {\em Data-Driven Modeling \& Scientific Computation: Methods for
  Complex Systems \& Big Data}.
\newblock Oxford University Press, Inc., USA, 2013.

\bibitem{Lafon-2004}
S.S. Lafon.
\newblock {\em Diffusion Maps and Geometric Harmonics}.
\newblock {PhD} dissertation, Yale University, 2004.

\bibitem{lee2020model}
Kookjin Lee and Kevin~T Carlberg.
\newblock Model reduction of dynamical systems on nonlinear manifolds using
  deep convolutional autoencoders.
\newblock {\em Journal of Computational Physics}, 404:108973, 2020.

\bibitem{Lehmberg2020}
Daniel Lehmberg, Felix Dietrich, Gerta K{\"o}ster, and Hans-Joachim Bungartz.
\newblock datafold: data-driven models for point clouds and time series on
  manifolds.
\newblock {\em Journal of Open Source Software}, 5(51):2283, 2020.

\bibitem{Limechanism}
Juan Li, Zhenwei Zhao, Andrei Kazakov, and Frederick~L Dryer.
\newblock An updated comprehensive kinetic model of hydrogen combustion.
\newblock {\em International journal of chemical kinetics}, 36(10):566--575,
  2004.

\bibitem{papaioannou2021time}
Panagiotis Papaioannou, Ronen Talmon, Ioannis Kevrekidis, and Constantinos
  Siettos.
\newblock Time series forecasting using manifold learning, 2021.

\bibitem{raissi2019physics}
Maziar Raissi, Paris Perdikaris, and George~E Karniadakis.
\newblock Physics-informed neural networks: A deep learning framework for
  solving forward and inverse problems involving nonlinear partial differential
  equations.
\newblock {\em Journal of Computational Physics}, 378:686--707, 2019.

\bibitem{rico1992discrete}
R~Rico-Martinez, K~Krischer, IG~Kevrekidis, MC~Kube, and JL~Hudson.
\newblock Discrete-vs. continuous-time nonlinear signal processing of cu
  electrodissolution data.
\newblock {\em Chemical Engineering Communications}, 118(1):25--48, 1992.

\bibitem{shlizerman2012proper}
Eli Shlizerman, Edwin Ding, Matthew~O Williams, and J~Nathan Kutz.
\newblock The proper orthogonal decomposition for dimensionality reduction in
  mode-locked lasers and optical systems.
\newblock {\em International Journal of Optics}, 2012, 2012.

\bibitem{shvartsman2000order}
Stanislav~Y Shvartsman, C~Theodoropoulos, Roberto Rico-Mart{\i}nez,
  IG~Kevrekidis, Edriss~S Titi, and TJ~Mountziaris.
\newblock Order reduction for nonlinear dynamic models of distributed reacting
  systems.
\newblock {\em Journal of Process Control}, 10(2-3):177--184, 2000.

\bibitem{sonday-2011}
B.~Sonday.
\newblock {\em Systematic Model Reduction for Complex Systems Through Data
  Mining and Dimensionality Reduction}.
\newblock {PhD} dissertation, Princeton University, 2011.

\bibitem{strogatz:2000}
Steven~H. Strogatz.
\newblock {\em Nonlinear Dynamics and Chaos: With Applications to Physics,
  Biology, Chemistry and Engineering}.
\newblock Westview Press, 2000.

\bibitem{Swift_Hohenberg}
Mario Taboada.
\newblock Finite-dimensional asymptotic behavior for the swift-hohenberg model
  of convection.
\newblock {\em Nonlinear Analysis: Theory, Methods \& Applications}, 14(1):43
  -- 54, 1990.

\bibitem{talmon2013diffusion}
Ronen Talmon, Israel Cohen, Sharon Gannot, and Ronald~R Coifman.
\newblock Diffusion maps for signal processing: A deeper look at
  manifold-learning techniques based on kernels and graphs.
\newblock {\em IEEE signal processing magazine}, 30(4):75--86, 2013.

\bibitem{Finite_Difference_Trefethen1996FiniteDA}
Lloyd~Nicholas Trefethen.
\newblock {\em Finite difference and spectral methods for ordinary and partial
  differential equations}.
\newblock Cornell University-Department of Computer Science and Center for
  Applied, 1996.

\bibitem{ZHANG2003429}
Yongchun Zhang, Michael~A. Henson, and Yannis~G. Kevrekidis.
\newblock Nonlinear model reduction for dynamic analysis of cell population
  models.
\newblock {\em Chemical Engineering Science}, 58(2):429--445, 2003.

\end{thebibliography}

\newpage
\appendix
\onecolumn
\section{Appendix}
\section{Methodology}
\label{sec:Methodology}

\subsection{Diffusion Maps}
\label{sec:Diffusion_Maps}
\paragraph{Dimensionality reduction}
In this use case, the Diffusion Maps algorithm can reveal the intrinsic geometry of a given data set $\textbf{X}$. This is achieved by first defining the local similarity between the sampled data points on $\textbf{X}$.
With this similarity, a weighted graph  $\mathbf{K} \in \mathbb{R}^{N \times N}$ is constructed between all data points. The entries of this graph are often computed with a kernel function. In our case, the diffusion kernel was used:
\begin{equation} 
\label{eq:Kernel}
{K}(\vect{x}_i,\vect{x}_j) = \exp \left (-\frac{|| \vect{x}_i - \vect{x}_j||_2^2}{2\epsilon} \right ),
\end{equation}
where $\epsilon$ is a hyper-parameter that specifies the (square of the) kernel bandwidth \cite{Gear_Chiavazzo_Parametrizing_boundary_detection}. If points $\vect{x}_i,\vect{x}_j$ satisfy \hbox{$|| \vect{x}_i - \vect{x}_j||_2^2 < \epsilon$}, they are considered similar (connected), whereas points with distance larger than $\epsilon^2$ are, effectively, not linked directly. Even though in this paper only the Euclidean distance of the points was used, in general, different metrics are also possible.

It is worth mentioning that Diffusion Maps is based on the approximation of the heat kernel $e^{-t\Delta}$, where $\Delta$ is the Laplace-Beltrami operator on a manifold $M$, in the limit $n \rightarrow \infty, \epsilon \rightarrow 0 $ \cite{COIFMAN20065,TimeScaleSeparation}. Because the sampling density affects such a kernel estimate, a normalization is needed to recover the correct operator:  
\begin{eqnarray}
    P_{ii} &=& \sum_{j=1}^{N} K_{ij},\\
        \widetilde{\textbf{K}} &=&\textbf{P}^{-\alpha}\textbf{K}\textbf{P}^{-\alpha}.
\end{eqnarray}
The bias parameter $\alpha$ is being introduced to control the influence that the sampling density has on the geometry. When $\alpha = 0$ is chosen the influence of density is maximal and we get the approximation of the Laplace-Beltrami operator only in the case of uniform sampling \cite{CoifmanTomographyLaplacian}. In the case $\alpha = 1$ the density effects are factored out and the Laplace-Beltrami operator approximation is obtained.
A second normalization is applied,
\begin{equation}
    {A}(\vect{x}_i,\vect{x}_j) = \frac{{\widetilde{K}}(\vect{x}_i,\vect{x}_j)}{\sum_{j=1}^{N} {\widetilde{K}}(\vect{x}_i,\vect{x}_j)},
\end{equation}
which leads to the construction of $\textbf{A}$, a row-stochastic, Markovian matrix. The eigendecomposition of $\textbf{A}$ has a set of eigenvectors, $ \vect{\phi}_i$ and eigenvalues $\lambda_{i}.$
\begin{equation}
    \label{eq:Eigendecomposition}
    \textbf{A}\vect{\phi}_i = \lambda{_i}\vect{\phi}_i
\end{equation}

Appropriate selection of the eigenvectors $\vect{\phi}_{i}, i \geq 1 ,$ that correspond to unique eigendirections yields a more parsimonious representation of the data \cite{dsilva2015parsimonious}. More precisely, if the number of those independent non-harmonic eigenvectors (e.g. $k$) is smaller than the dimensions of the ambient space ($k < m$) of $\textbf{X}$  those $k$ new coordinates, $\mathbf{\Phi}= \{\vect{\phi}_1,\vect{\phi}_2, ..., \vect{\phi}_k \}$, can be used for a new lower dimensional parametrization of the original high dimensional data set.

\subsection{Harmonic and Non-Harmonic Eigenvectors}
\label{sec:Harmonics_Non_Harmonics}
This section aims to provide a more clear theoretical justification of our techniques and discuss the selection of the \textit{non-harmonic} Diffusion Maps coordinates. We start by reminding the reader how the selection of the important \textit{modes} is performed with Proper Orthogonal Decomposition (P.O.D) to give a more clear comparison with the selection with Diffusion Maps. In P.O.D.  one firstly samples the behavior of the underlying dynamical system to obtain a representative data ensemble. Next, one applies Singular Value Decomposition (S.V.D.) to this data set. The spectrum of the singular values computed gives an indication of how many dimensions are necessary to capture the important \textit{modes} of a dynamical system. This is based on the assumption that the leading singular vectors that capture most of the variance capture also the most important dynamics. A useful reduced dimensionality of a given data set with P.O.D. is chosen as a compromise between reduction of complexity and reconstruction error. From here, expressing the reduced dynamics can be done by performing a projection of the original dynamics to the chosen important modes \cite{deane1991low,Kuhn_Data_Driven_Modeling_Scientific_Computations}.

For Diffusion Maps the selection of the latent coordinates that parametrize the data set should not be done based (only) on the information provided by the eigenvalues even though in some case an eigengap can give a hint \cite{talmon2013diffusion}. The reason is that some eigenvectors are parametrizing the same direction. We refer to those eigenvectors, as \textit{harmonic} eigenvectors and for the set of eigenvectors that parametrize different directions as \textit{non-harmonic}. Therefore our goal of discovering the minimal set of Diffusion Maps eigenvectors is restricted on selecting the \textit{non-harmonic} eigenvectors. 

\begin{figure}[h]
    \begin{center}
    \includegraphics[scale=0.45]{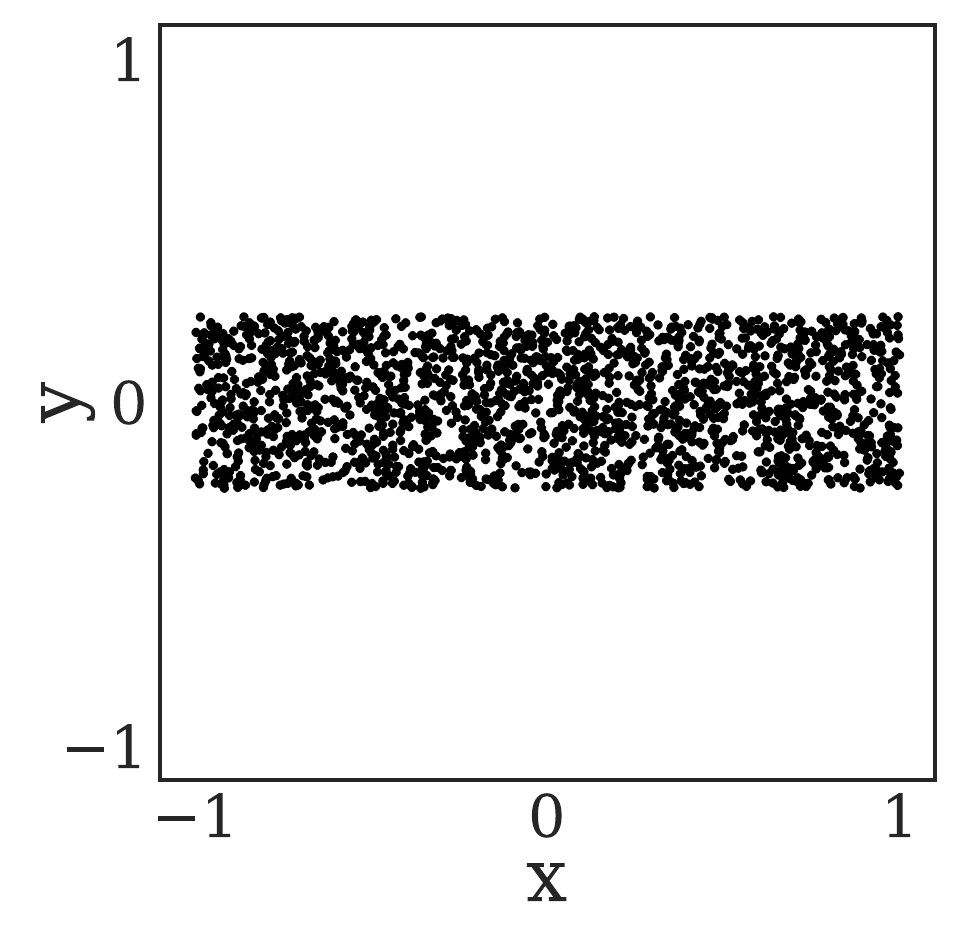}
    \caption{Data points collected from rectangle with sides' ratio 4:1.}
    \label{fig:rectangle}
    \end{center}
    \end{figure}

In the remaining of this section we provide a toy example to elucidate the concepts of \textit{harmonic} and \textit{non-harmonic} eigenvectors . We sampled uniformly 10,000 data points on a rectangle with ratio of it's length to width equal to 4:1, Figure \ref{fig:rectangle}. For this toy example since the Diffusion Maps algorithm (for $\alpha=1$ or for data with uniform sampling density) approximates the Laplace-Beltrami operator with Neumann boundary conditions the eigenvalues and the eigenvectors can be computed also analytically  \cite{chen2019selecting,dsilva2015parsimonious} and it is known that the order of the Diffusion Maps coordinates that parametrize this data set is based on the ratio of it's sides \cite{chen2019selecting}.

\begin{figure}[h]
    \begin{center}
    \includegraphics[scale=0.45]{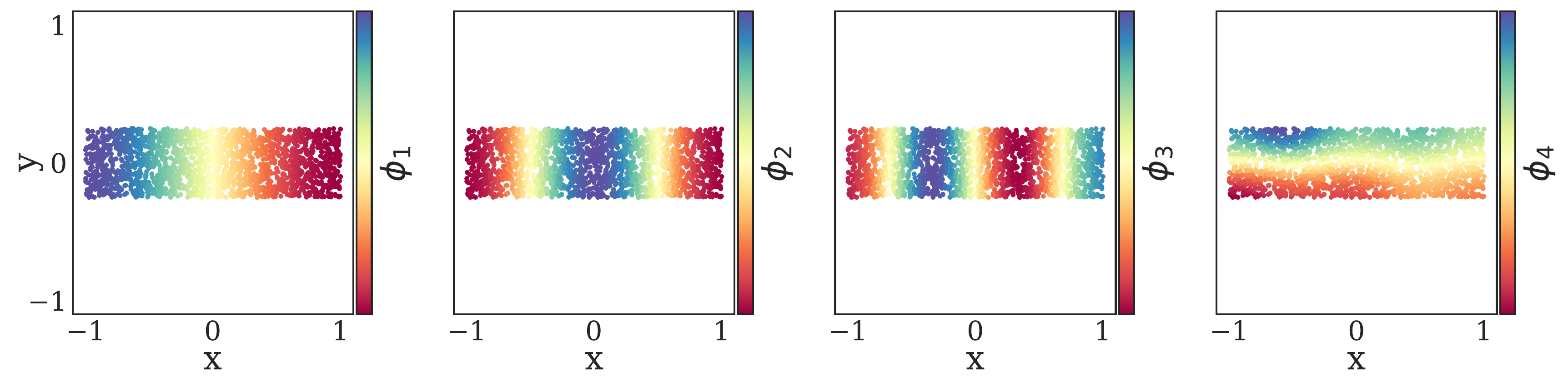}
    \caption{The first four (non-trivial) Diffusion Maps coordinates are colored as function on the data set.}
    \label{fig:rectangle_colored_dmap}
    \end{center}
    \end{figure}

We performed Diffusion Maps on the sampled data points and rank the eigenvectors based on their eigenvalues. The first Diffusion Maps eigenvector $\vect{\phi}_0$ with $\lambda_0 =1 $ is the \textit{trivial} constant eigenvector therefore is discarded. In Figure \ref{fig:rectangle_colored_dmap} we color the first four (non-trivial) Diffusion Maps coordinates as functions on the data set. As it can be seen from Figure \ref{fig:rectangle_colored_dmap} $\phi_1$ parametrize the same direction as the $x$ coordinate; and is actually one to one with $x$. The next two Diffusion Maps coordinates, $\vect{\phi}_2$ and $\vect{\phi}_3$, parametrize also the same direction as $\phi_1$ in different ``frequencies'', therefore they are harmonics of $\vect{\phi}_1$. However, $\vect{\phi}_4$ appears to parametrize the $y$ direction (is one to-one) and thus is non-harmonic to $\vect{\phi}_1$.

\begin{figure}[h]
    \begin{center}
    \includegraphics[scale=0.45]{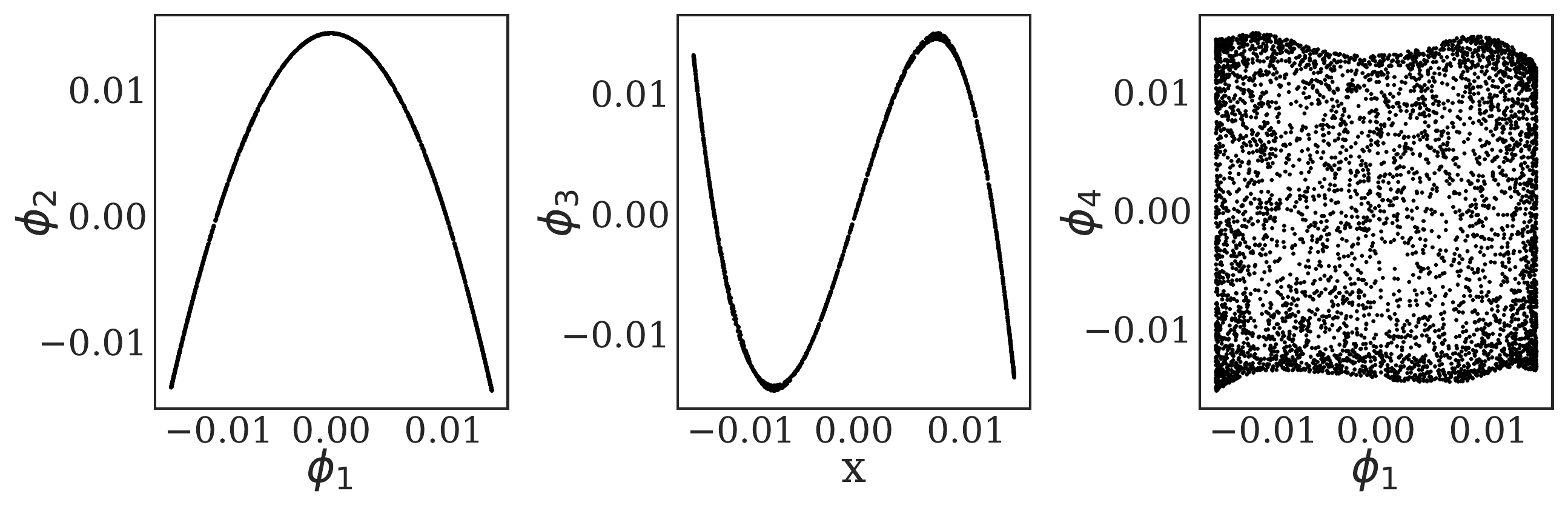}
    \caption{The first non-trivial coordinate $\vect{\phi}_1$ is plotted as against the eigenvectors $\vect{\phi}_2$, $\vect{\phi}_3$, $\vect{\phi}_4$ respectively, .}
    \label{fig:phis_versus}
    \end{center}
    \end{figure}

In our illustrated toy example the two non-harmonic eigenvectors parametrize independently two of the original coordinates of the data set (similar to the Chafee-Infante example discussed in Section \ref{sec:Chafee_Infante}); this won't be the case in any given data set. In many cases the Diffusion Maps coordinates will parametrize combinations (linear or nonlinear) of the ambient data set's coordinates. In those cases a visual selection of the non-harmonic eigenvectors is also possible if the number of non-harmonic eigenvectors is smaller or equal to three. In these cases, the selection starts by plotting the leading  non-trivial Diffusion Maps coordinate $\vect{\phi}_1$ versus the eigenvectors with lower values of eigenvalues. From Figure \ref{fig:phis_versus}, $\vect{\phi}_2$ and $\vect{\phi}_3$ can be seen as functions of $\vect{\phi}_1$. On the contrary, the direction that $\vect{\phi}_4$ spans is independent of $\vect{\phi}_1$ since given a value of $\vect{\phi}_1$ there is a range of  possible values for $\vect{\phi}_4$. In a data set where we might suspect also a third dimension constructing three dimensional visualizations of the candidate eigenvectors in terms of the first two non-harmonic eigenvectors could be performed.

In higher dimensions where visual selection of the non-harmonic eigenvectors is not feasible algorithms that compute the independent coordinates become possible. Carmeline et al. \cite{dsilva2015parsimonious} proposed an automated algorithm that detects coordinates corresponding to harmonic eigenvectors based on local linear regression. For our examples illustrated in Sections \ref{sec:Chafee_Infante} and \ref{sec:Combustion} we selected the latent coordinates with visual inspection and as a sanity check by the local linear regression algorithm \cite{dsilva2015parsimonious}.

\subsubsection{Nystr\"om Extension}
\label{sec:Nystrom Extension}
The Nystr\"om method is a technique for finding numerical approximations to eigenfunction problems of the form \cite{EfficientspatiotemporalNystrom}
\begin{equation}
    \label{eq:NystromIntegral}
    \int_{a}^b {K}(\vect{x}_i,\vect{x}_j)\vect{\phi}({\vect{x}_j) = \lambda\vect{\phi}(\vect{x_i}}).
\end{equation}
In our work, the Nystr\"om extension is mainly used to generate a Diffusion Map coordinate $\vect{\phi}_{new}$ for a sample point $\vect{x}_{new} \notin \vect{X}$. This interpolation function, $f:\vect{x}_{new} \rightarrow \vect{\phi}_{new}$, is being obtained by evaluating the kernel that was used during the dimensionality reduction step (and apply the same normalizations) on the new point $\vect{x}_{new}$. More precisely, the Euclidean distance between this new point, $\vect{x}_{new}$ and all the preexisting points in the data set $\mathbf{X}$ is computed as:
\begin{equation} 
\label{eq:Kernel_nystrom}
{K}(\vect{x}_{new},\vect{x}_j) = \exp \left (-\frac{|| \vect{x}_{new} - \vect{x}_j||^2}{2\epsilon} \right ),
\end{equation}

\begin{equation}
     {\widetilde{K}}(\vect{x}_{new},\vect{x}_j) = \frac{{K}(\vect{x}_{new},\vect{x}_j)}{{p}(\vect{x}_{new})^{\alpha}\vect{p}(\vect{x}_j)^{\alpha}},
\end{equation}

with $p(\vect{x}_{new}) = \sum_{j=1}^N {{K}}(\vect{x}_{new},\vect{x}_j)$ being just a scalar value and  $\vect{p}(\vect{x}_{i}) = \sum_{j=1}^N {K}(\vect{x}_i,\vect{x}_j)$ being a N-dimensional vector. Then,
\begin{equation}
    {A}(\vect{x}_{new},\vect{x}_i) = \frac{{\widetilde{K}}(\vect{x}_{new},\vect{x}_j)}{\sum_{j = 1}^N {\widetilde{K}}(\vect{x}_{new},\vect{x}_j)},
\end{equation}
and the value of the reduced coordinate $\beta^{th}$ is computed through
\begin{equation}
\label{eq:Nystrom_Expression}
    {\phi}_{\beta} (\vect{x}_{new}) = \frac{1}{\lambda_{\beta}} \sum_{i=1}^N {A}(\vect{x}_{new},\vect{x}_i)\phi_{\beta}(x_{i}),
\end{equation}
where $\phi_{\beta}(x_{i})$ is the $i$-th component of the $\beta$-th eigenvector ($\vect{\phi}_{\beta}$) and $\lambda_{\beta}$ is the $\beta$-th eigenvalue. 
Since the Nystr\"om Extension provides a mapping from the ambient space coordinates $\vect{x}$ to the reduced coordinates $\vect{\phi}$, we can also get a closed form expression of derivative of the Nystr\"om Extension.
Symbolic differentiation of Equation \eqref{eq:Nystrom_Expression} for $\alpha =0$ gives:

\begin{multline}
        \label{eq:Nystrom_derivative}
        \frac{\partial\phi_{\beta}}{\partial{x_{new,k}}} = \frac{1}{\lambda_{\beta}} \frac{\partial}{\partial{x_{new,k}}} \left(\sum_{i=1}^N {A}(\vect{x}_{new},\vect{x}_j)\phi_{j,\beta} \right) = \\ {\frac{1}{\lambda_{\beta}} \frac{\sum_{i,j = 1}^N  K(\vect{x}_{new},\vect{x}_j)\left(\frac{\partial}{\partial{x_{new,k}}}K(\vect{x}_{new},\vect{x}_i) \right) (\phi_{i,{\beta}} -\phi_{j,{\beta}})}{\left (\sum_{i = 1}^N K(\vect{x}_{new},{\vect{x}_i}) \right)^2}}
\end{multline}
where,
\begin{equation}
    \label{eq:Nystrom_derivative2}
    \frac{\partial}{\partial{x_{new,k}}}K(\vect{x}_{new},\vect{x}_i) = 2\epsilon^{-2}K(\vect{x}_{new},{\vect{x}_i})(x_{i,k} -x_{new,\beta}).
\end{equation}
This derivative can then be used to map vectors in the ambient space (e.g., time derivatives) to the latent space.
\clearpage
\section{Equation-free projective integration} 
\label{sec:Equation_Free}
Similar to the aforementioned BF scheme is the \textit{Equation-free} approach. In this paper we describe this particular scheme without using it for our examples; a more thorough explanation of this algorithm can be found in \cite{kevrekidis2003equation}. In the Equation Free approach, sampling of the entire manifold is not needed in advance. Sampling the manifold ``locally'' is enough in order to initiate this scheme. In this sampled data set Diffusion Maps can be performed in order to get a set of reduced latent coordinates. On those coordinates we need to estimate the gradients. This can be done by computing numerically the derivative with finite differences in terms of the reduced coordinates and perform \textit{projective integration}. Alternatively, as suggested in \cite{koelle2018manifold} instead of estimating the gradients by using differences between nearby points $\phi(x_i), \phi(x_{i+1}) $ one can estimate their values {\em after first projecting the points to the tangent space} of $\phi(x_i)$. In general a (projective) integration step leads to a new point in the reduced coordinates \textit{outside} the sampled manifold. Given this out of sample point \textit{lifting} in the ambient space coordinates can be achieved with Latent Harmonics (Algorithm \ref{alg:DDMaps}). For this new coordinate in the ambient space, where closed form expressions of the dynamics are available, sampling again the manifold locally and repeating this procedure is possible.
\clearpage
\section{Additional Results}
\label{sec:Additional_Results}
\subsection{Chafee-Infante}
\begin{figure}[ht]
\vskip 0.2in
\begin{center}
\includegraphics[scale=0.35]{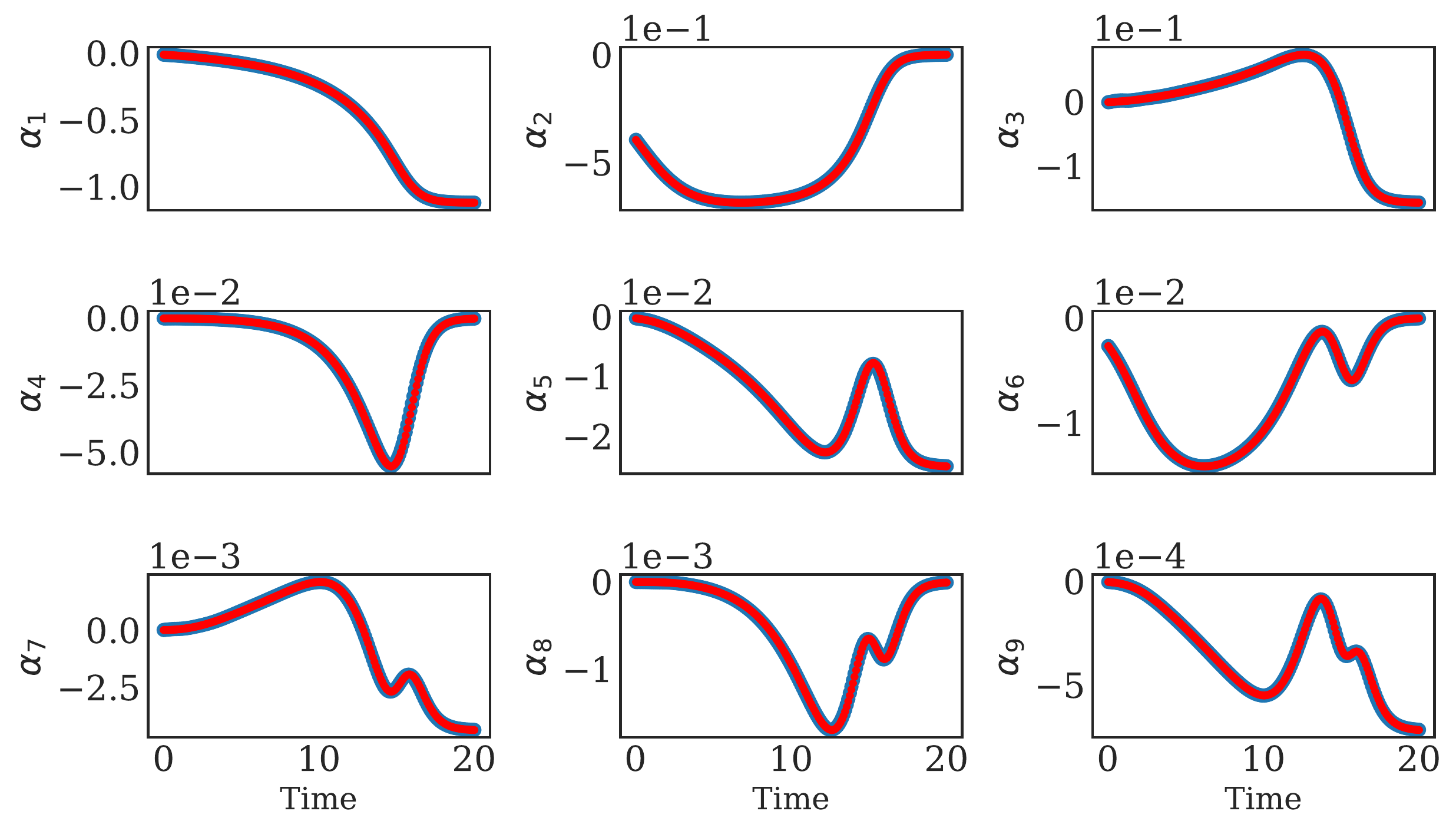}
\caption{Comparison  between the lifted with Latent Harmonics trajectory computed with the data  driven  integration scheme Back and Forth (blue trajectory)  and  the trajectory coming from the spectral odes (red trajectory).}
\label{fig:CI_DMAPs_High_Dimensional_Back_and_Forth}
\end{center}
\vskip -0.2in
\end{figure}

\begin{figure}[ht]
\vskip 0.2in
\begin{center}
\includegraphics[scale=0.35]{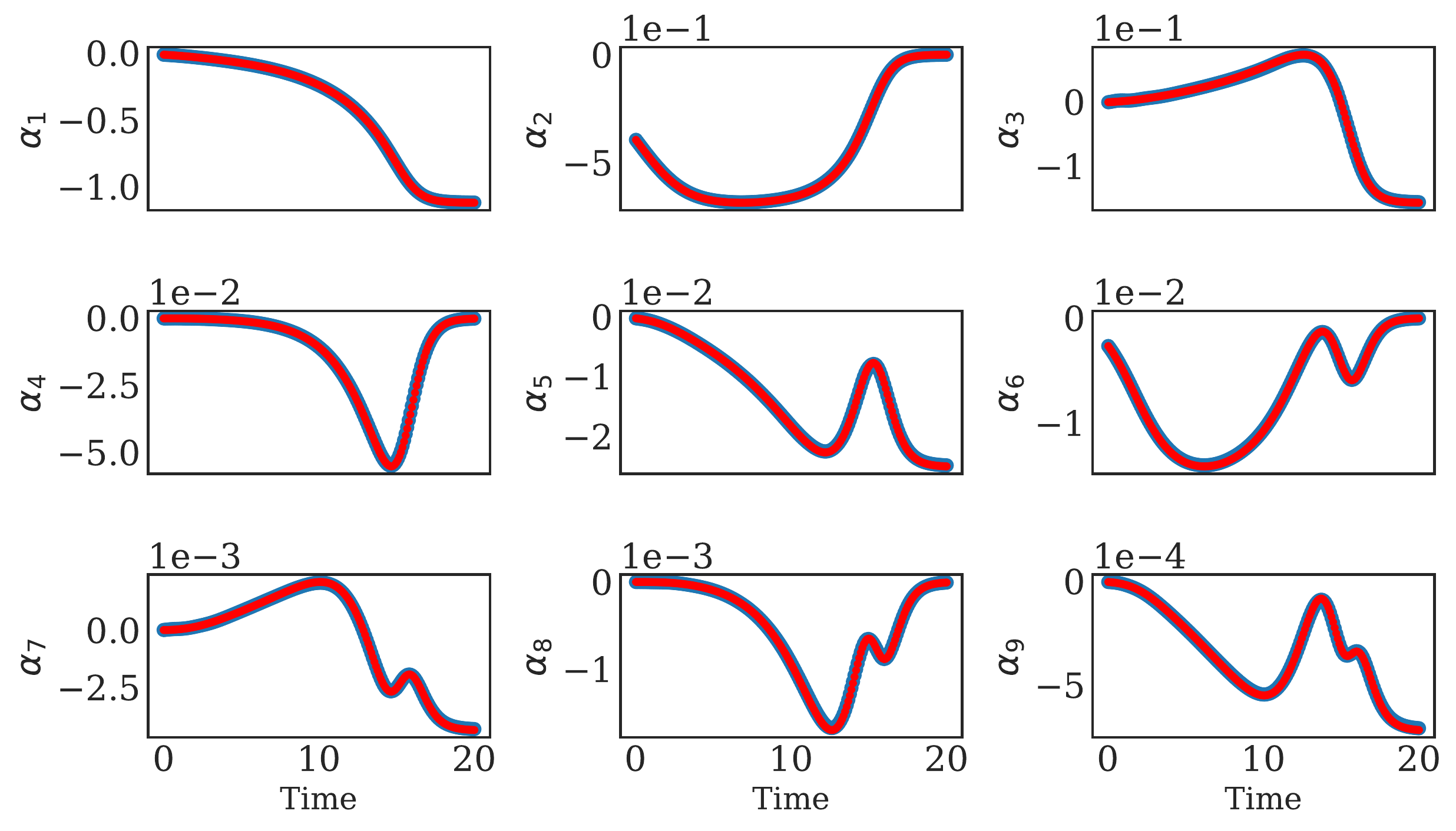}
\caption{Comparison  between the lifted with Latent Harmonics trajectory computed with the data-driven  integration  scheme Grid Tabulation (blue trajectory)  and  the trajectory coming from the spectral odes (red trajectory)}
\label{fig:CI_DMAPs_High_Dimensional_GT}
\end{center}
\vskip -0.2in
\end{figure}

A qualitative comparison is given between the lifted trajectories computed with the data-driven suggested approaches (BF and GT) and the integrated spectral ODEs. Figure \ref{fig:CI_DMAPs_High_Dimensional_Back_and_Forth} shows the comparison for the BF approach and Figure \ref{fig:CI_DMAPs_High_Dimensional_GT} shows the comparison for the GT approach.

\clearpage
\subsection{Chemical Kinetics}
\begin{figure}[ht]
\vskip 0.2in
\begin{center}
\includegraphics[scale=0.35]{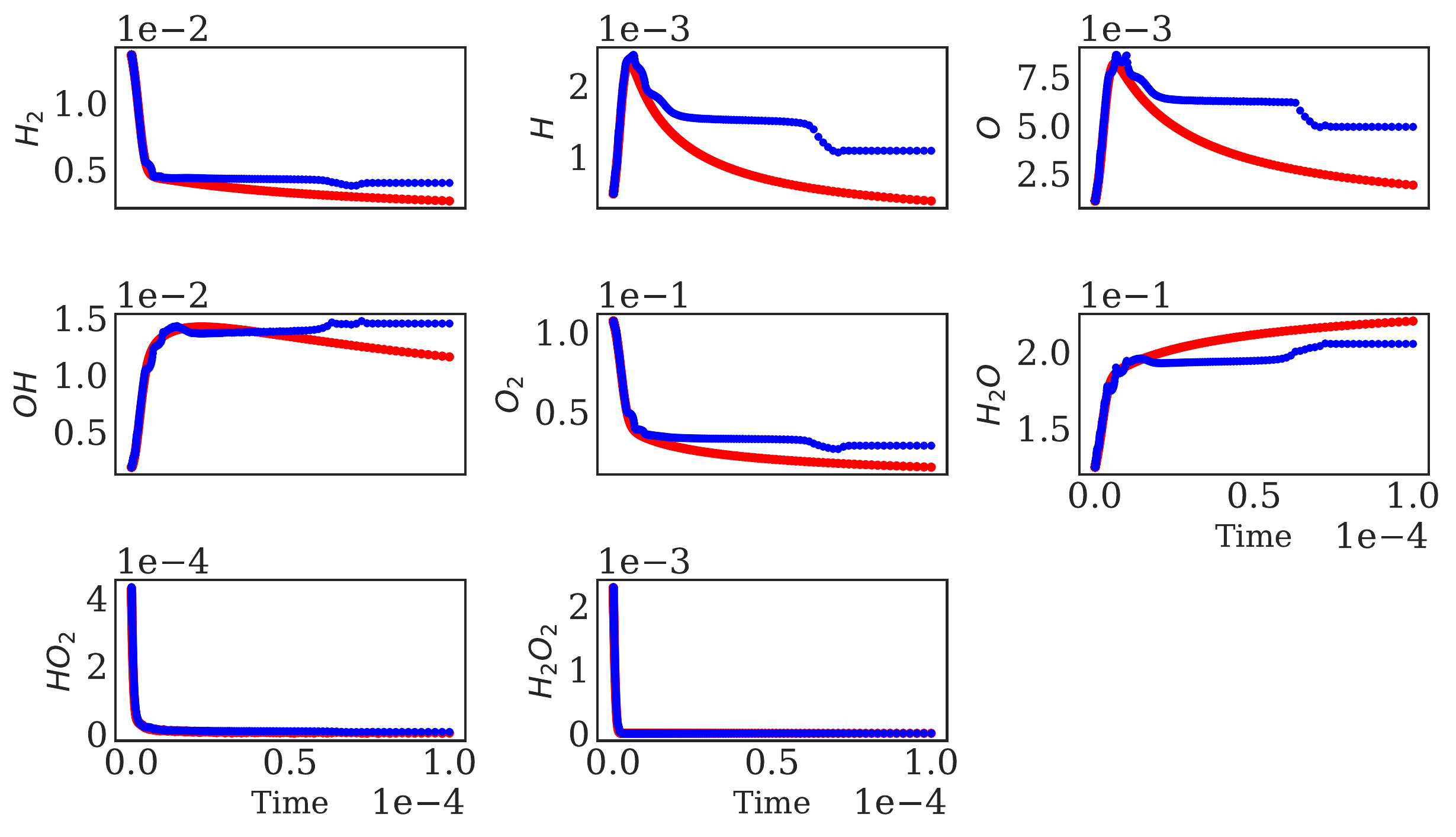}
\caption{Comparison  between the lifted with Latent Harmonics trajectory computed with the data-driven  integration  scheme Back and Forth (blue trajectory) and  the trajectory computed from the full, unreduced equations of the kinetics (red trajectory).}
\label{fig:CO_DMAPs_High_Dimensional_BF}
\end{center}
\vskip -0.2in
\end{figure}

\begin{figure}[ht]
\vskip 0.2in
\begin{center} 
\includegraphics[scale=0.35]{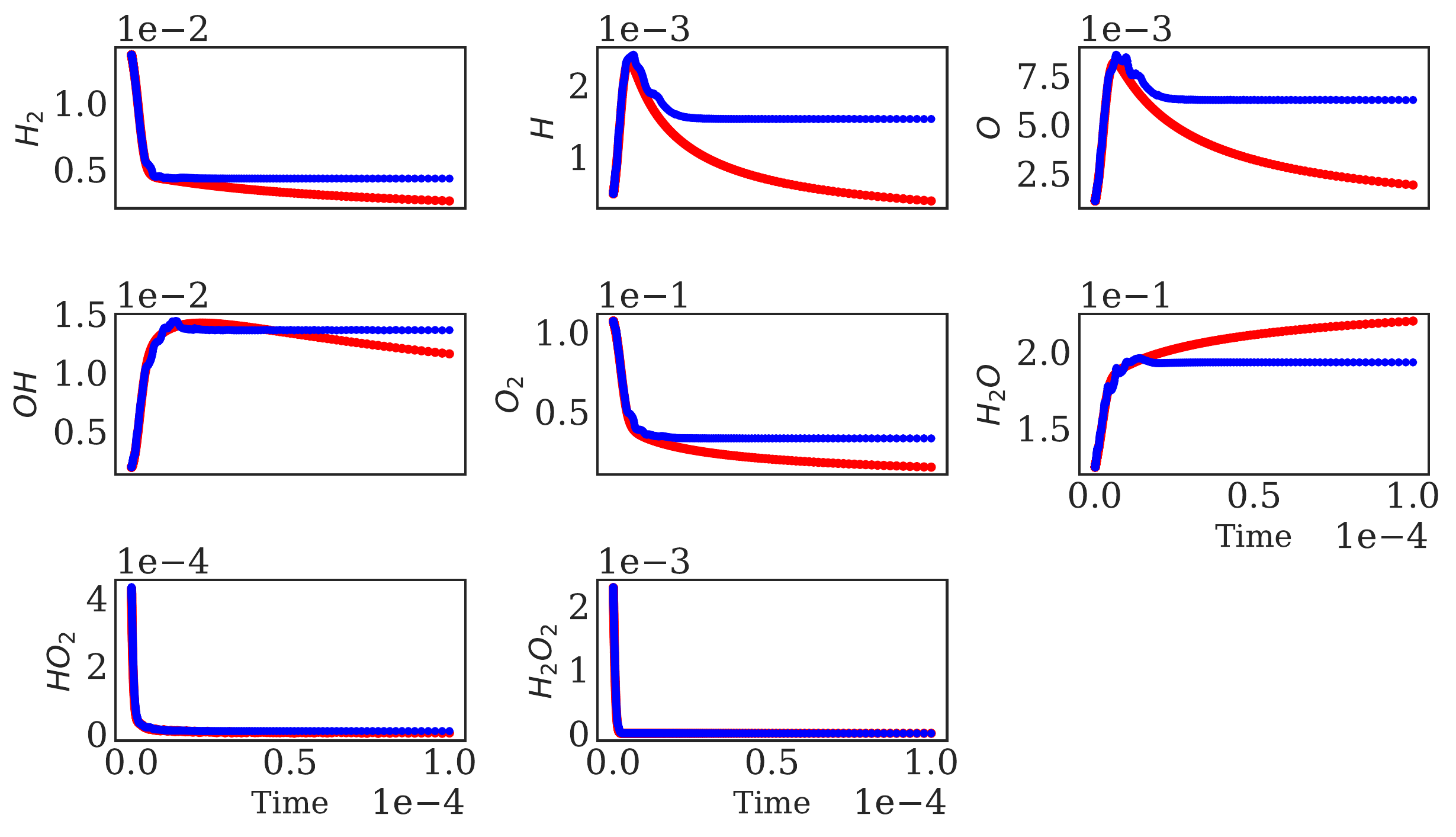}
\caption{Comparison  between the lifted with Latent Harmonics trajectory computed with the data-driven  integration  scheme Grid Tabulation (blue trajectory) and the trajectory computed from the full, unreduced equations of the mechanism (red trajectory). The reduced trajectory due to numerical errors is not able to continue since it \textit{leaves} the manifold.}
\label{fig:CO_DMAPs_High_Dimensional_GT}
\end{center}
\vskip -0.2in
\end{figure}

For the Chemical Kinetics example the qualitative comparison is given between the lifted trajectories computed with the data-driven suggested approaches (BF and GT) and the integrated system of ODEs. Figure \ref{fig:CO_DMAPs_High_Dimensional_BF} shows the comparison for the BF approach and Figure \ref{fig:CO_DMAPs_High_Dimensional_GT} shows the comparison for the GT approach.

\clearpage
\section{Diffusion Maps and Latent Harmonics Parameters Selection}
\label{sec:DMaps_specifics}
On Tables \ref{tab:Tableparameters1} and \ref{tab:Tableparameters2} we report the different parameters used for Diffusion Maps and also for Latent Harmonics for the two examples.

\begin{table}[h]
    \centering
    \caption{Parameters used for Diffusion Maps and Latent Harmonics for the Chafee-Infante PDE}
    \begin{tabular}{|c|c|c|c|c|c|}
        \hline 
         Method & $\epsilon$ & \# eigenvalues & cut off &  normalization ($\alpha$)  & Markov Matrix \\ \hline\hline
        Diffusion Maps & $ 0.584 $ &  9  & $\infty$ &  0 & Yes \\ \hline 
        Latent Harmonics & $ 4.46 \times 10^{-05}$ & 300  & $\infty$ &  0 & No  \\ \hline
    \end{tabular}

    \label{tab:Tableparameters1}
\end{table}

\begin{table}[h]
    \centering
    \caption{Parameters used for Diffusion Maps and Latent Harmonics for the Combustion of Hydrogen.}
    \begin{tabular}{|c|c|c|c|c|c|}
        \hline 
         Method & $\epsilon$ & \# eigenvalues & cut off &  normalization ($\alpha$)  & Markov Matrix \\ \hline\hline
        Diffusion Maps & $ 18 $ &  9  & $\infty$ &  0 & Yes \\ \hline 
        Latent Harmonics & $ 4.11 \times 10^{-05}$ & 400  & $\infty$ &  0 & No  \\ \hline
    \end{tabular}

    \label{tab:Tableparameters2}
\end{table}
\clearpage
\section{Errors}
\label{sec:errors}
\begin{figure}[ht]
\vskip 0.2in
\begin{center}
\includegraphics[scale=0.32]{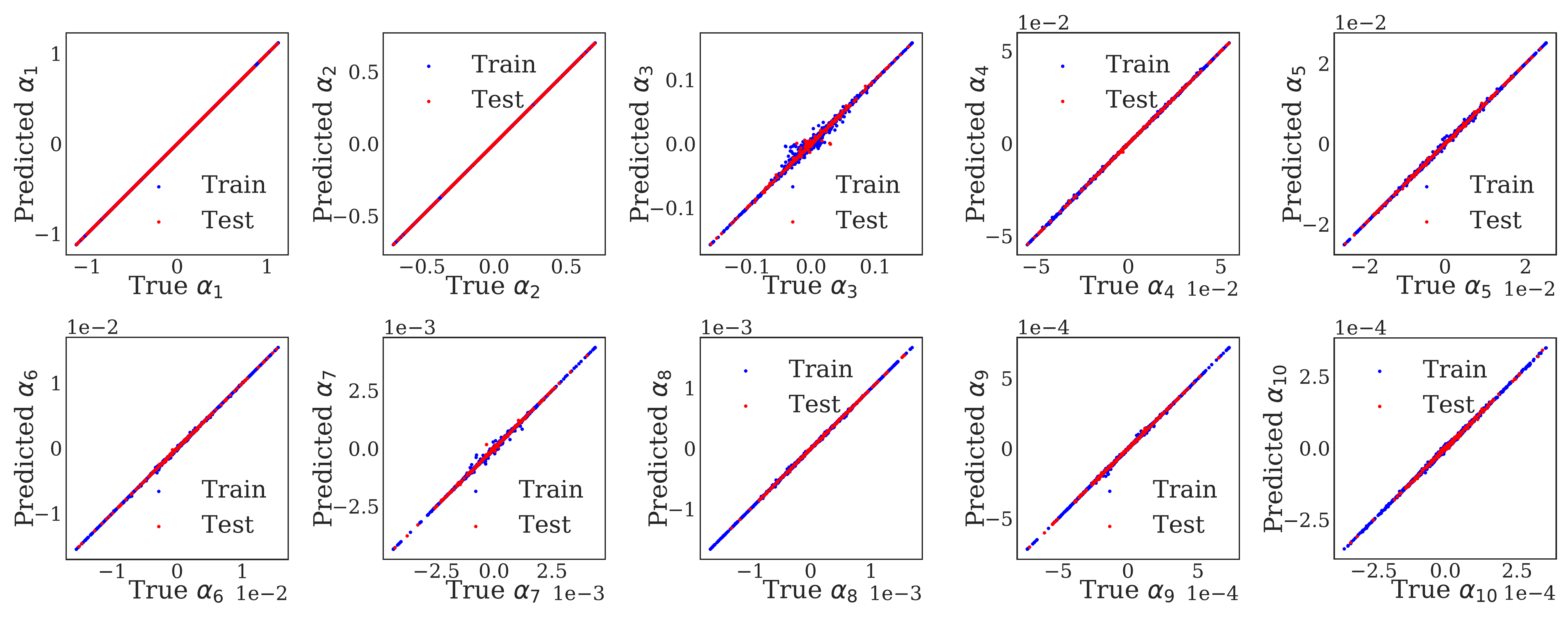}
\includegraphics[scale=0.37]{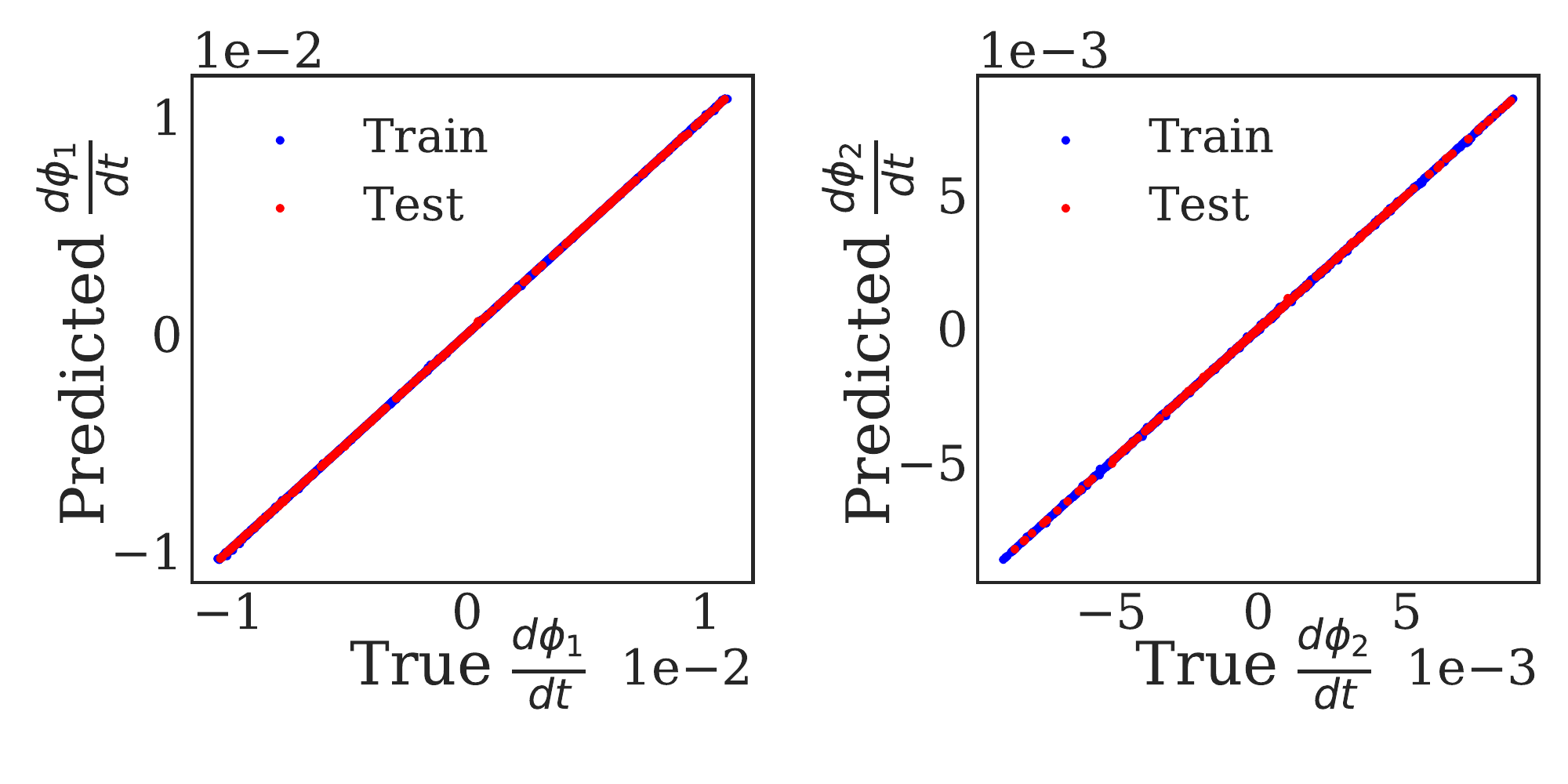}
\caption{The true values of functions of interest plotted against the predicted with Latent Harmonics for the Chafee Infante PDE. Blue color points represent the training points and red points the test.}
\label{fig:errors-lifting-CI}
\end{center}
\vskip -0.2in
\end{figure}

\begin{figure}[ht]
\centering 
\includegraphics[scale=0.32]{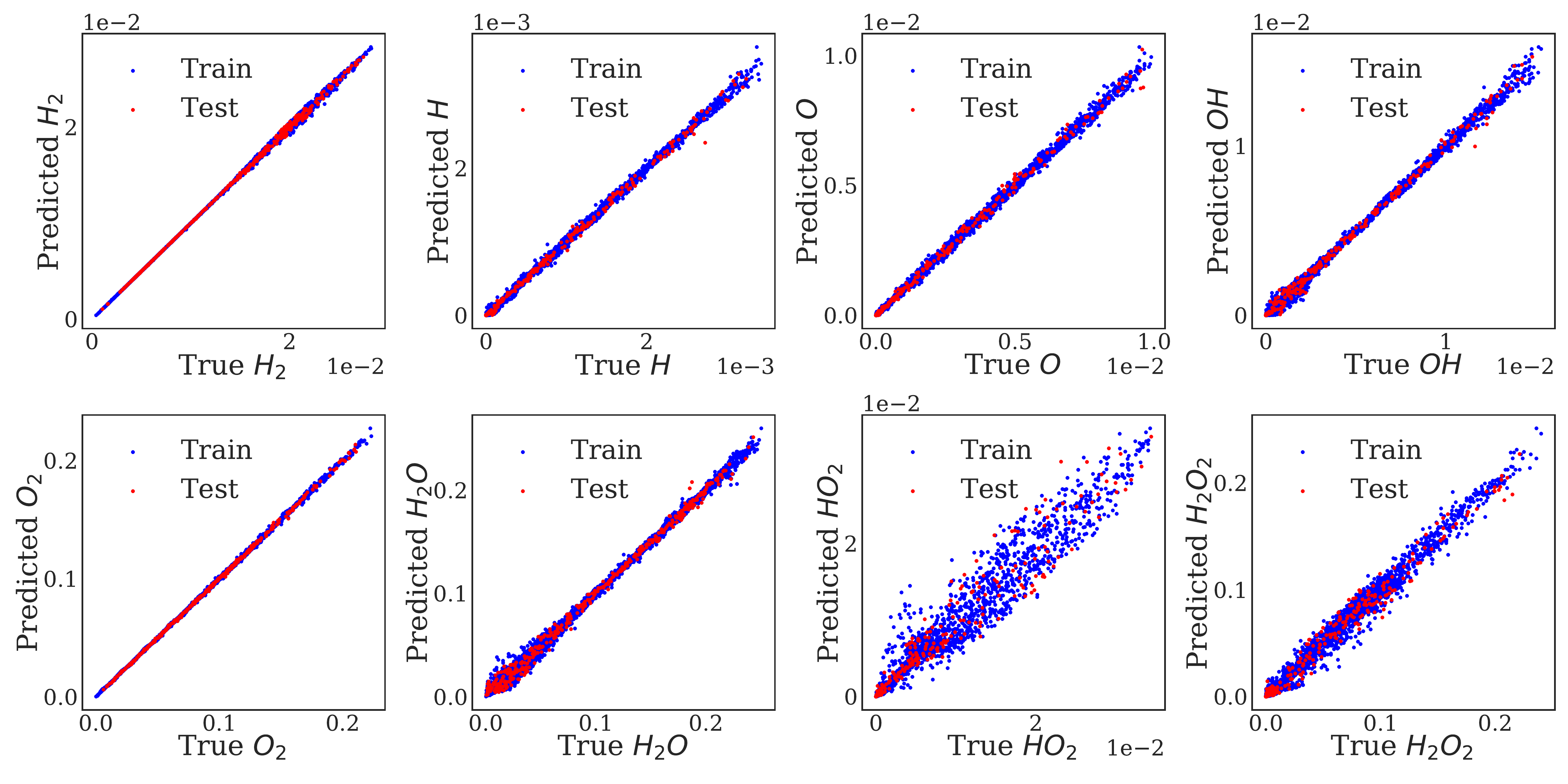}
\includegraphics[scale=0.38]{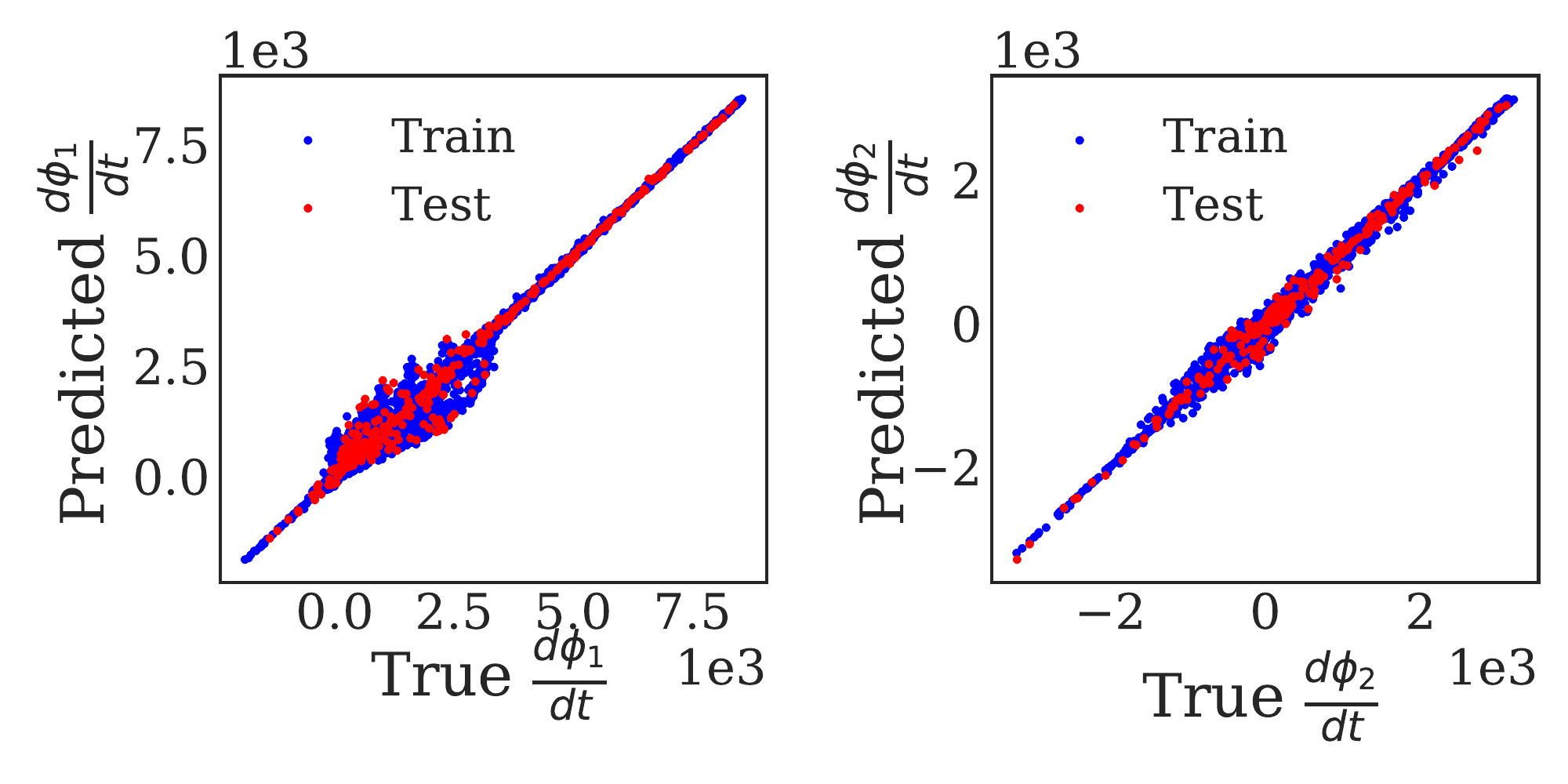}
\caption{The true values of functions of interest plotted against the predicted with Latent Harmonics for the Kinetics example. Blue color points represent the training points and red points the test.}
\label{fig:errors-lifting-CO}
\end{figure}

In this section we report the Mean Squared Errors for our interpolation schemes with Latent Harmonics. The reported errors in Tables \ref{tab:TableMSE1} and \ref{tab:TableMSE2} are computed on test data sets that contain 10\% of the total sampled data points.  Before the use of our interpolation scheme the functions of interest were scaled and translated  in the range zero to one by using the \textit{MinMaxScaler} from scipy.
However, in Figures \ref{fig:errors-lifting-CI} and \ref{fig:errors-lifting-CO} the variables are being transformed back to their original values.
\subsection*{Chafee-Infante Example - Errors}
\begin{table}[h]
     \caption{The Mean Squared Error of the functions of interest were computed on a test set for the Chafee-Infante PDE}
    \begin{tabular}{|c|c|c|c|c|c|}
        \hline 
          & $\alpha_1$ & $\alpha_2$ & $\alpha_3$ &  $\alpha_4$  & $\alpha_5$  \\ \hline\hline
        Mean Squared Error &  1.41 $\times 10^{-9}$  &  1.10 $\times 10^{-10}$  & 2.39 $\times 10^{-6}$ &   7.59$\times 10^{-6}$ & 2.68 $\times 10^{-6}$ \\ \hline
    \end{tabular}
        \begin{tabular}{|c|c|c|c|c|c|}
        \hline 
            & $\alpha_6$ & $\alpha_7$ & $\alpha_8$ &  $\alpha_9$  & $\alpha_{10}$ \\ \hline\hline
        Mean Squared Error & 2.68 $\times 10^{-6}$ & 1.65 $\times 10^{-5}$ & 1.67 $\times 10^{-6}$ & 4.37 $\times 10^{-6}$ & 1.64 $\times 10^{-5}$  \\ \hline
    \end{tabular}
    \centering
        \begin{tabular}{|c|c|c|}
        \hline 
            & $ \frac{d\phi_1}{dt}$ & $ \frac{d\phi_2}{dt}$  \\ \hline\hline
        Mean Squared Error & 3.40 $\times 10^{-7}$ & 7.84 $\times 10^{-7}$   \\ \hline
    \end{tabular}

         \label{tab:TableMSE1}
\end{table}

\clearpage
\subsection*{Chemical Kinetics - Errors}

\begin{table}[ht]
     \caption{The Mean Squared Error of the functions of interest were computed on a test set for the Chemical Kinetics example.}
    \begin{tabular}{|c|c|c|c|c|}
        \hline 
          & $\ce{H2}$ & $\ce{H}$ & $\ce{O}$ &  $\ce{OH}$   \\ \hline\hline
        Mean Squared Error &  1.51 $\times 10^{-5}$  &  5.11 $\times 10^{-2}$  & 8.41 $\times 10^{-2}$ &   5.79$\times 10^{-2}$  \\ \hline
    \end{tabular}
        \begin{tabular}{|c|c|c|c|c|}
        \hline 
            & $\ce{O2}$ & $\ce{H2O}$ & $\ce{HO2}$ &  $\ce{H2O2}$   \\ \hline\hline
        Mean Squared Error & 2.41 $\times 10^{-5}$ & 2.24 $\times 10^{-2}$ & 2.05 $\times 10^{-2}$ & 1.17 $\times 10^{-1}$  \\ \hline
    \end{tabular}
    \centering
        \begin{tabular}{|c|c|c|}
        \hline 
            & $ \frac{d\phi_1}{dt}$ & $ \frac{d\phi_2}{dt}$  \\ \hline\hline
        Mean Squared Error & 9.07 $\times 10^{-4}$ & 2.27 $\times 10^{-4}$   \\ \hline
    \end{tabular}

         \label{tab:TableMSE2}
\end{table}

\clearpage
\section{Codes}
All the codes that generate the figures and construct the suggested algorithms will be uploaded on a public repository on Gitlab upon acceptance.

\end{document}